\pdfoutput=1

\documentclass[11pt]{article}

\usepackage{acl}[final]

\usepackage{times}
\usepackage{latexsym}

\usepackage[T1]{fontenc}

\usepackage[utf8]{inputenc}

\usepackage{microtype}

\usepackage{inconsolata}

\usepackage{multirow}
\usepackage{makecell}
\usepackage[justification=justified,skip=5pt]{caption} %
\usepackage{subcaption}
\usepackage{graphicx}
\usepackage[inline]{enumitem} 
\usepackage{makecell}
\usepackage{soul} 
\usepackage{color, xcolor} 
\usepackage{booktabs}
\usepackage{mdframed}
\usepackage{amsmath}
\usepackage{tabularx}
\usepackage{adjustbox}
\usepackage{amssymb}
\usepackage{bbding}
\usepackage{pifont}
\usepackage{cleveref}
\usepackage{tikz}
\usepackage{tikzpagenodes}
\usepackage{longtable}
\usetikzlibrary{arrows.meta}
\usetikzlibrary{backgrounds}
\usetikzlibrary{positioning, fit, mindmap, trees, calc, tikzmark, shapes}
\usetikzlibrary{shapes.arrows, fadings, automata, tikzmark, shapes.geometric, decorations.pathreplacing, patterns}
\usetikzlibrary{arrows.meta}
\usetikzlibrary{matrix}
\usetikzlibrary{intersections}
\usetikzlibrary{tikzmark}
\usepackage{tkz-euclide}
\usepackage{tikzpeople}
\usepackage{sistyle}
\SIthousandsep{,}
\usepackage{url}
\definecolor{warm orange}{HTML}{f46d43}
\definecolor{cool blue}{HTML}{74add1}
\DeclareMathOperator{\similarity}{sim}

\makeatletter
\g@addto@macro\normalsize{%
  \abovedisplayskip 4pt plus 2pt minus 3pt%
  \belowdisplayskip \abovedisplayskip
  \abovedisplayshortskip 4pt plus2pt  minus3pt%
  \belowdisplayshortskip 4pt plus2pt minus3pt%
}

\makeatother

\title{Blinded by Generated Contexts: How Language Models\\ Merge Generated and Retrieved Contexts When Knowledge Conflicts?}

  \author{Hexiang Tan$^{\spadesuit\heartsuit}$, Fei Sun$^{\spadesuit}\footnotemark[2]
  $, Wanli Yang$^{\spadesuit}$, Yuanzhuo Wang$^{\spadesuit}$, Qi Cao$^{\spadesuit}$, Xueqi Cheng$^{\spadesuit\heartsuit}$ \\
  $^{\spadesuit}$CAS Key Laboratory of AI Safety,\\
  Institute of Computing Technology, Chinese Academy of Sciences, Beijing, China \\
  $^{\heartsuit}$University of Chinese Academy of Sciences, Beijing, China \\
  \tt{\{tanhexiang21s, sunfei, wangyuanzhuo, caoqi, cxq\}@ict.ac.cn}}
  
\begin{document}
\maketitle

\renewcommand*{\thefootnote}{\fnsymbol{footnote}}
\footnotetext[2]{Corresponding author.}
\renewcommand*{\thefootnote}{\arabic{footnote}}

\begin{abstract}
While auxiliary information has become a key to enhancing Large Language Models (LLMs), relatively little is known about how LLMs merge these contexts, specifically contexts generated by LLMs and those retrieved from external sources.
To investigate this, we formulate a systematic framework to identify whether LLMs' responses are attributed to either generated or retrieved contexts.
To easily trace the origin of the response, we construct datasets with conflicting contexts, i.e., each question is paired with both generated and retrieved contexts, yet only one of them contains the correct answer.
Our experiments reveal a significant bias in several LLMs (GPT-4/3.5 and Llama2) to favor generated contexts, even when they provide incorrect information.
We further identify two key factors contributing to this bias: i) contexts generated by LLMs typically show greater similarity to the questions, increasing their likelihood of being selected; ii) the segmentation process used in retrieved contexts disrupts their completeness, thereby hindering their full utilization in LLMs.
Our analysis enhances the understanding of how LLMs merge diverse contexts, offers valuable insights for advancing current LLM augmentation methods, and highlights the risk of generated misinformation for retrieval-augmented LLMs\footnote{Code released at \url{https://github.com/Tan-Hexiang/RetrieveOrGenerated}.}.

\end{abstract}

\section{Introduction}

Recent advancements in augmenting Large Language Models (LLMs) with auxiliary information have significantly revolutionized their efficacy in knowledge-intensive tasks \cite{chang2023survey,ram2023context}.
This auxiliary information can originate from contexts generated by LLMs or retrieved from external sources.
For the former, \citet{liu2021generated, sun2022recitation, yu2022genread} instruct LLMs to initially generate a background context tailored to the given question, which then serves as the basis for producing the final answer.
In contrast, retrieval-augmented approaches \citep{lewis2020retrieval,ram2023context} include relevant passages retrieved from external corpora, such as Wikipedia, as context, thereby notably enhancing LLMs' capability to address challenges like knowledge updates \cite{jang2022temporalwiki} and long-tail knowledge \cite{kandpal2023longtail}.

\begin{figure}[t]
    \centering
    \includegraphics[width=0.92\linewidth]{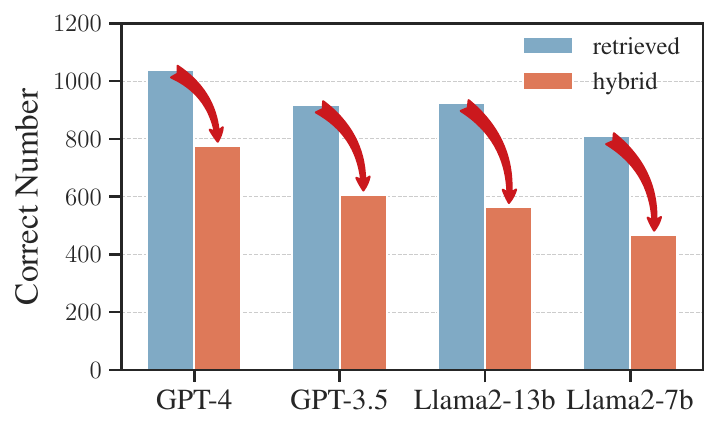}
    \caption{\textcolor{cool blue}{Blue} bars show the success number on the NQ test set with only retrieved contexts, while \textcolor{warm orange}{orange} bars depict the decline in success for the same questions when generated contexts are additionally incorporated.}
    \label{figure_introduction}
\end{figure}

Recent hybrid methods have attempted to integrate these two types of contexts to further improve performance in tasks like Question Answering (QA) \citep{yu2022genread, mallen2022when, zhang2023merging}.
However, we observe an abnormal phenomenon: in certain cases, models relying solely on retrieval contexts succeeded, whereas hybrid approaches failed, as depicted in \Cref{figure_introduction}.
This observation implies that LLMs may struggle to effectively integrate diverse types of contexts, overlooking correct information in retrieved contexts.
To uncover the reasoning behind this, this study investigates \textit{the underlying mechanisms by which LLMs process the two types of contexts, especially when they contain conflicting information.}

To facilitate the investigation, we proposed a systematic framework to dissect the process by which LLMs merge generated and retrieved contexts.
We curated tailored \textbf{c}ontext-\textbf{c}onflicting (CC) datasets in which each question is accompanied by a pair of generated and retrieved contexts.
These contexts are deliberately designed to be inconsistent, with only one containing the correct answer to its corresponding question.
These datasets provide a solid foundation for determining whether LLMs utilize retrieved or generated context to produce responses in QA tasks.

In this paper, we conducted a series of controlled experiments using our uniquely designed datasets to empirically study this question, focusing on several state-of-the-art closed (GPT-3.5/4) and open (Llama2-7b/13b) LLMs.
Surprisingly, our findings reveal a pronounced bias in LLMs to \textit{favor generated contexts} across various LLMs (\S\ref{subsection_Other-Generated Contexts}), even when the generated contexts offer incorrect information while the retrieval contexts hold the correct answers.
Further analysis shows that the bias is prevalent across various retrieval models (\S\ref{sec 4.3}).
These findings highlight a critical challenge for existing LLMs in effectively merging contexts from diverse sources, especially in light of the increasing prevalence of LLM-generated content on the internet, which may contain potential misinformation \citep{pan2023misinformation, chen2023combating}.

Through extensive empirical analyses, we reveal several factors contributing to this bias:
\begin{enumerate*}[label=(\roman*)]
    \item
    \textit{confirmation bias \citep{xie2023knowledgeconflicts} is not a key factor} (\S\ref{sec 5.1}): LLMs maintain a significant preference for generated contexts even when they are inconsistent with LLMs' parametric knowledge.
    \item \textit{text similarity is a significant factor} (\S\ref{section_similarity}): compared to retrieved contexts, generated contexts typically exhibit a higher degree of similarity to the questions, even when they contain incorrect information.  
    The samples with a larger similarity gap between generated and retrieved contexts exhibit a more pronounced bias. 
    \item \textit{semantic completeness matters} (\S\ref{section_completeness}): 
    LLMs tend to favor contexts with semantic integrity.
    The segmentation process used in retrieved contexts may disrupt their completeness, thereby hindering their full utilization in LLMs.
    This finding emphasizes the need for optimizing passage segmentation in current retrieval systems.
\end{enumerate*}

In summary, we explore the challenges LLMs face when utilizing conflicting contexts and make the following contributions:
\begin{itemize}[leftmargin=11pt, itemsep=3pt, topsep=0pt, partopsep=0pt, parsep=0pt]
    \item We uncover a critical bias in existing LLMs, where they heavily rely on generated contexts regardless of correctness, indicating an insufficient use of diverse information sources.
    \item To facilitate controlled experiments, we develop a specialized framework for constructing tailored datasets and excluding confounding factors, e.g. input order, and context length.
    \item 
    Our extensive analyses have identified two key factors, i.e., text similarity and semantic completeness, in the context utilization of LLMs. 
\end{itemize}

\section{Background \& Study Formulation}
\begin{figure*}[t]
    \centering
    \begin{minipage}{0.49\textwidth}
        \begin{minipage}{\textwidth}
            \centering
            \includegraphics[width=\textwidth]{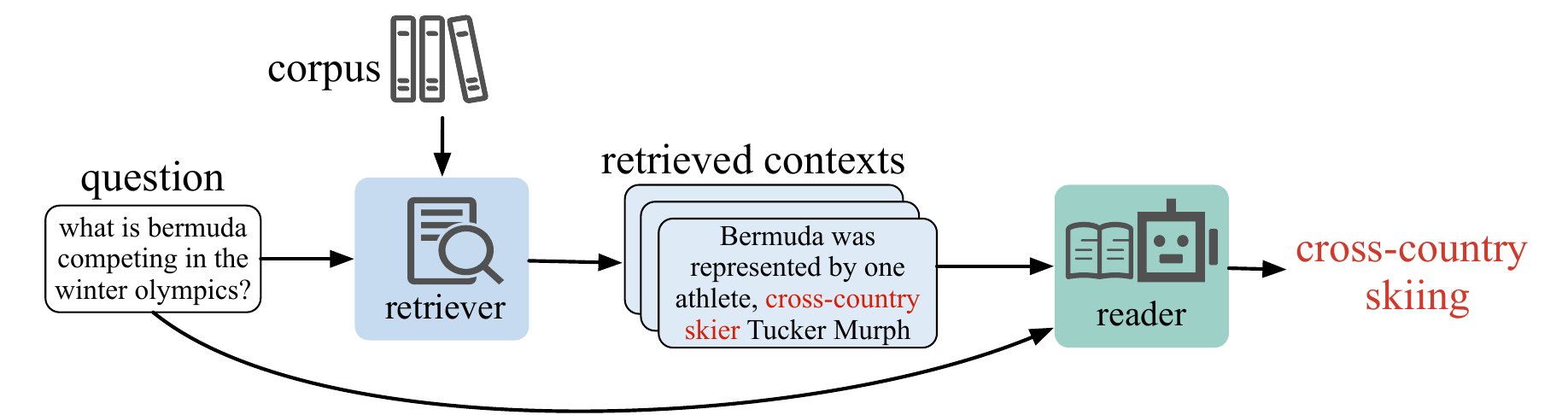}
            \subcaption{Retrieval-Augmented Approach}
            \label{pipeline RAG}
        \end{minipage} \\
        \begin{minipage}{\textwidth}
            \centering
            \includegraphics[width=\textwidth]{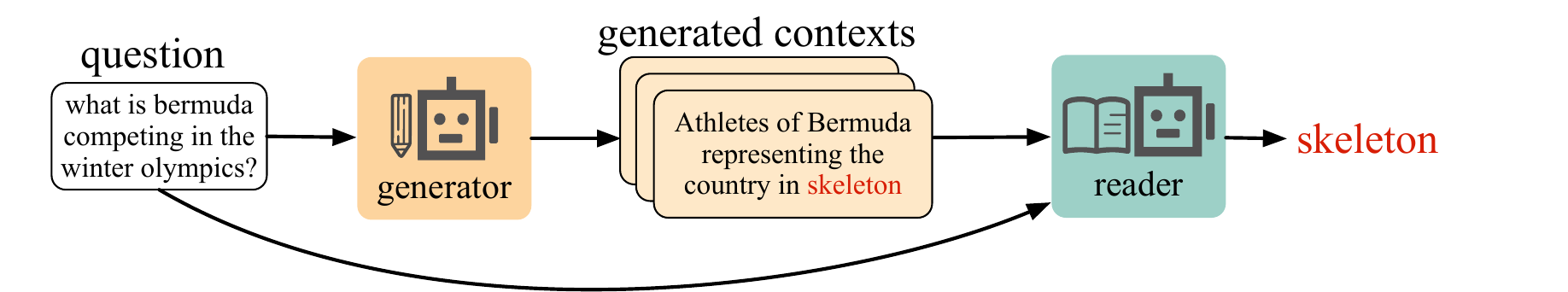}
            \subcaption{Generation-Augmented Approach}
            \label{pipeline GAG}
        \end{minipage}
    \end{minipage}
    \begin{minipage}{0.49\textwidth}
        \centering
        \includegraphics[width=\textwidth]{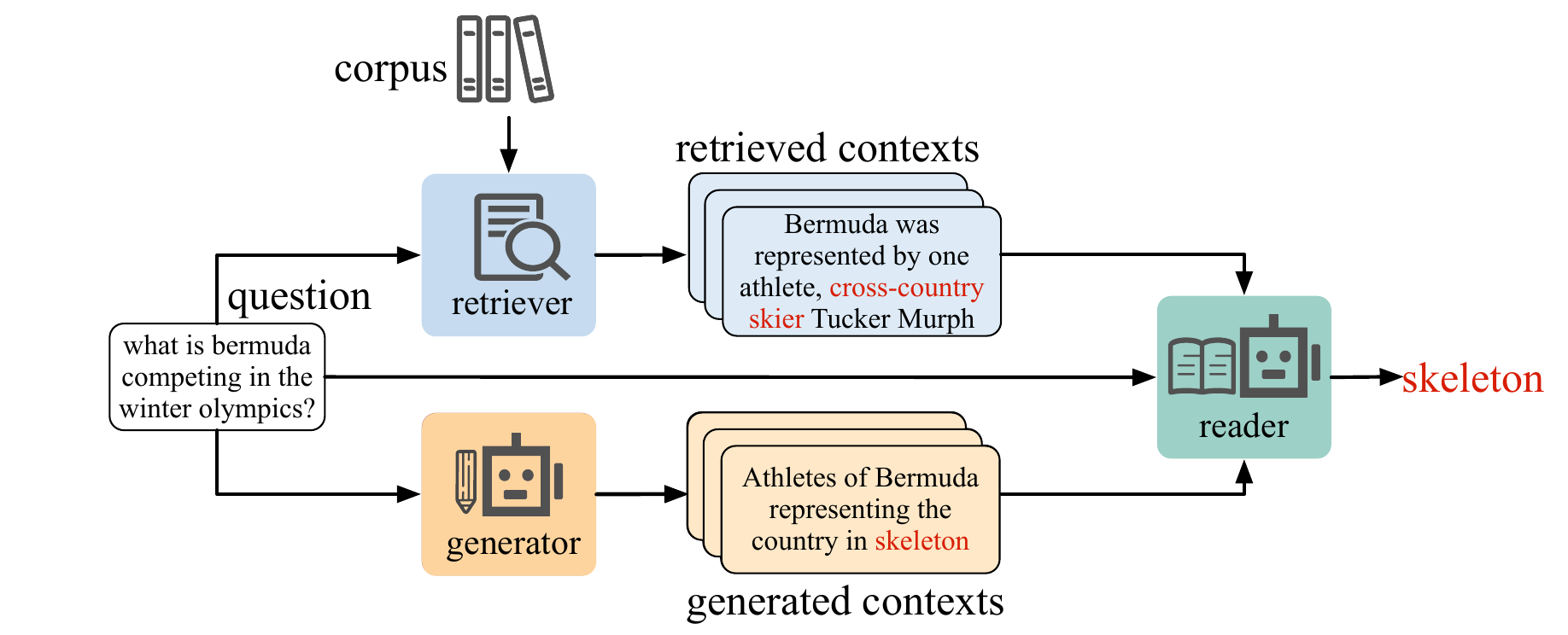}
        \subcaption{Hybrid Approach}
        \label{pipeline Hybrid Approach}
    \end{minipage}
    \caption{The frameworks of retrieval-augmented approach, generation-augmented approach, and hybrid approach.}
    \label{pipeline}
\end{figure*}

In this section, we briefly review three categories of LLMs augmented with auxiliary information for QA tasks: retrieval-augmented, generation-augmented, and hybrid approaches.
Additionally, we introduce the framework of our investigation.
\subsection{Background}
\Cref{pipeline} presents high-level abstract frameworks for three typical types of QA systems, each centered around an LLM as the \textit{reader} component, and potentially incorporating additional components like a \textit{retriever}, \textit{generator}, or a blend of both, tailored to the specific methodology.

\noindent\textit{\textbf{Retrieval-Augmented Approach}}.
As shown in \Cref{pipeline RAG}, for a given question $q$ in a set of questions $\mathbb{Q}$, these approaches \citep{guu2020realm,lewis2020retrieval,ram2023context,gao2023retrieval_survey} initially use a retrieval model $\gamma$ to select the top $k$ relevant documents $D_k^{\gamma} {=} \gamma_k(q, \mathbb{C}) {=} \{d_1^{\gamma}, \dots, d_k^{\gamma}\}$ from a corpus $\mathbb{C} {=} \{d_1,\dots, d_{|\mathbb{C}|}\}$.
Then, a reader (often LLM) $\phi$ employs these documents $D^{\gamma}_k$ to generate an answer $a_{\phi}^{\gamma}$, expressed as $a_{\phi}^{\gamma}=\phi(q,D^{\gamma}_{k})$.

\noindent\textit{\textbf{Generation-Augmented Approach}}.
In contrast, as illustrated in \Cref{pipeline GAG}, these works \citep{yu2022genread,sun2022recitation,liu2021generated} involve an LLM as a generator $\varrho$ to produce $k$ tailored background contexts $D_k^{\varrho} {=} \varrho_k(q) {=} \{d_1^{\varrho}, \dots, d_k^{\varrho}\}$ for a give question $q$, thereby enhancing the utilization of the LLM's internal knowledge.
These LLM-generated contexts $D^{\varrho}_k$ form the input for reader $\phi$ to produce the final answer: $a_{\phi}^{\varrho}=\phi(q, D^{\varrho}_k)$.

\noindent\textit{\textbf{Hybrid Approach}}, 
as depicted in \Cref{pipeline Hybrid Approach}, combines retrieved and generated contexts to enhance performance \cite{yu2022genread,abdallah2023generator}, as $a_{\phi}=\phi(q, D_k^{\gamma}, D_k^{\varrho})$.
These hybrid approaches face a significant challenge: conflicts between diverse sources can impede the effectiveness of information integration \cite{zhang2023merging}.

\noindent\textit{\textbf{Knowledge Conflicts within Contexts}}.
These studies mainly focus on conflicts within a \textit{single} type of input contexts, either only retrieved \citep{chen2022rich} or generated \citep{xie2023knowledgeconflicts}, leaving underexplored how LLMs resolve conflicts between diverse contexts.

\subsection{Answer Tracing Task}
\label{sec:task_def}

\begin{figure}[t]
    \centering
    \includegraphics[width=0.95\linewidth]{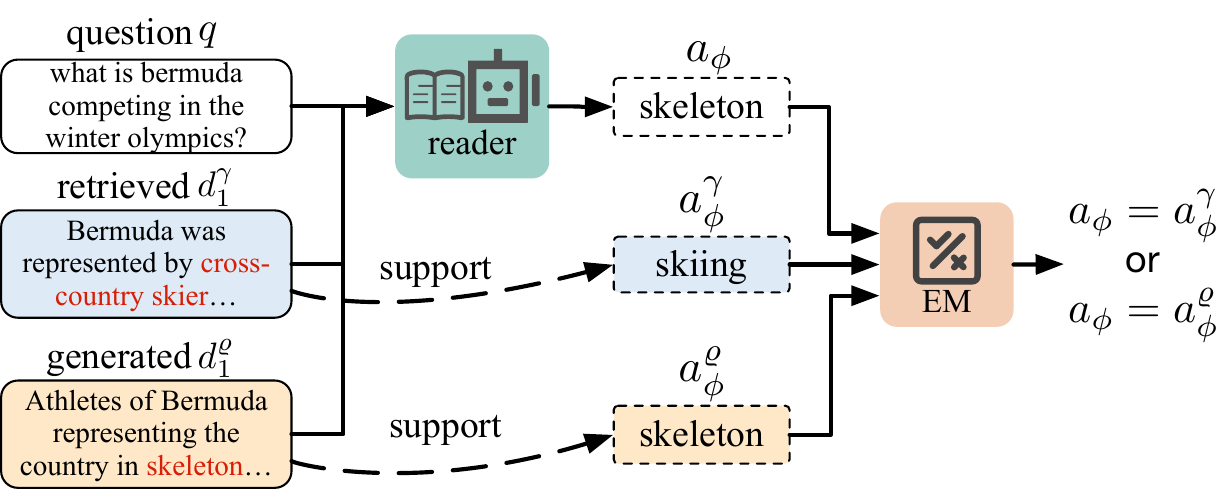}
    \caption{The task to study LLMs’ merging mechanisms
by tracing the sources of the answers.}
    \label{fig:task_formulation}
\end{figure}

Departing from previous research, our study investigates the mechanisms by which LLMs merge contexts from diverse sources in hybrid approaches. 
As illustrated in \Cref{fig:task_formulation}, we design a task to ascertain whether an answer $a_{\phi}$ originates from generated contexts $D_k^{\varrho}$ or retrieved contexts $D_k^{\gamma}$.
For a more controlled and simpler analysis, we limit the context to a single instance from each source, i.e., $k{=}1$ and $a_{\phi}{=}\phi(q, d^{\gamma}_{1}, d_1^{\varrho})$.
Then, by comparing the answer $a_{\phi}$ with the answers derived from the retrieved context $a_{\phi}^{\gamma}$ and the generated context $a_{\phi}^{\varrho}$, we can determine its source, thereby analyzing the merging mechanism of LLMs.

We specifically focus on non-tunable LLMs, i.e. in zero-shot settings, to reflect prevalent real-world use cases like ChatGPT.
This direction is motivated by the high cost and limited accessibility of fine-tuning, which makes the direct use of non-tunable LLMs popular.  
Additionally, given the extensive use of LLMs, any bias or issue in their merging mechanisms could lead to serious consequences.

\section{Experimental Setup} 
\label{Experimental Setup}

To facilitate our investigation into how LLMs merge generated and retrieved contexts, this section elaborates on the construction of our context-conflicting datasets and the evaluation metric. 

\subsection{Context-Conflicting  Datasets} \label{subsets}

\begin{figure*}[t]
    \centering
        \includegraphics[width=0.85\linewidth]{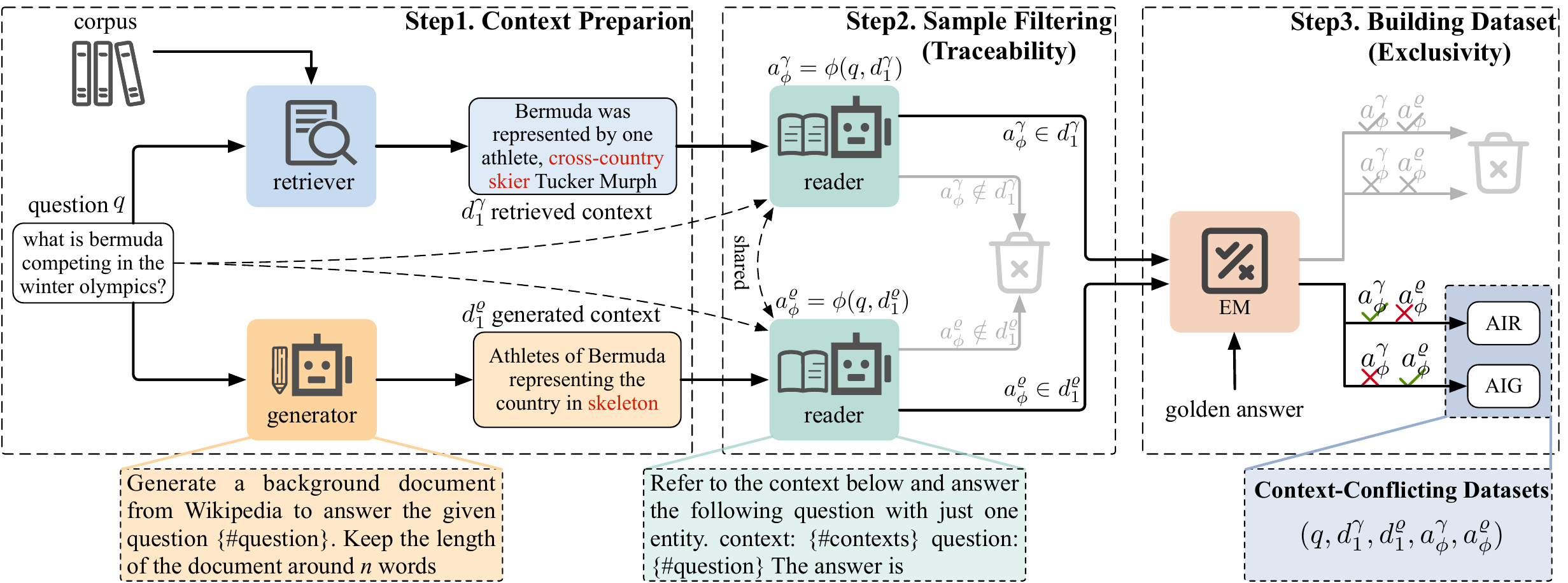}
        \label{fig:Datasets}
    \caption{The framework of constructing context-conflicting datasets.}
    \label{fig_dataset_construction}
\end{figure*}

As depicted in Figure \ref{fig_dataset_construction}, in our dataset $\mathcal{D}_{\mathrm{cc}}$, each sample $x$ is a quintet $(q, d_1^{\gamma}, d_1^{\varrho}, a_{\phi}^{\gamma}, a_{\phi}^{\varrho})$, where $d_1^{\gamma}$ is the context returned by retriever $\gamma$ for question $q$, $d_1^{\varrho}$ represents the context generated by LLM $\varrho$, $a_{\phi}^{\gamma}$ and $a_{\phi}^{\varrho}$ are the candidate answers provided by the reader $\phi$, each based solely on the respective contexts $d_1^{\gamma}$ and $d_1^{\varrho}$.
To guarantee that our dataset is suitable for controlled experiments aimed at investigating the merging mechanisms of LLMs, it should adhere to specific criteria:
\begin{itemize} [itemsep=3pt, topsep=0pt, partopsep=0pt, parsep=2pt, leftmargin=10pt] %
    \item \textbf{Traceability}: $a_{\phi}^{\gamma}$ and $a_{\phi}^{\varrho}$ should be supported by their corresponding contexts, $d_1^{\gamma}$ and $d_1^{\varrho}$.
    \item \textbf{Exclusivity}: Only one of the contexts, $d_1^{\gamma}$ or $d_1^{\varrho}$,  provides the correct answer, i.e., either $a_{\phi}^{\gamma}$ or $a_{\phi}^{\varrho}$ matches the gold answer of question $q$.
\end{itemize}
Such constraints establish a solid basis to identify which context, generated or retrieved, is selected by LLMs to produce answers in hybrid approaches.

We utilize the dev and test sets of two open-domain QA benchmark datasets with golden answers, i.e., NaturalQuestions (NQ) \cite{kwiatkowski2019natural} and TriviaQA (TQA) \cite{joshi2017triviaqa}, to assemble our experimental datasets.
The overall pipeline for dataset construction is depicted in \Cref{fig_dataset_construction}, with detailed steps outlined as follows:

\noindent\textbf{\textit{Context Preparation}}.
Step 1 in \Cref{fig_dataset_construction} illustrates the process of preparing contexts for each question.
For retrieved contexts, it is obtained from the top-$1$ ranked passage from Wikipedia using Contriever \citep{izacard2021unsupervised}, a powerful off-the-shelf retrieval model that is extensively employed in various retrieval-augmented generation systems \cite{shi2023replug,ram2023context}.

For generated context, we follow the GenRead \cite{yu2022genread}, instructing the generator, e.g., LLM like GPT-4, to generate a background document based on the question.
All LLMs in this paper, unless otherwise noted, have a temperature setting of zero to ensure result reproducibility.
However, this method often yields contexts much longer (${>}250$ words) than the retrieved contexts (typically truncated to ${\sim}100$ words \citep{dpr,izacard2021unsupervised}). The discrepancy in length could potentially affect the merging mechanisms of LLMs \citep{xie2023knowledgeconflicts}.
To exclude this disturbance, we regulate the length of the generated context by incorporating length constraint in the prompt, resulting in an average length discrepancy below $3$\%. 
All subsequent experiments, unless otherwise specified, employ this method to eliminate the impact of length variations.
More details can be found in Appendix \ref{appendix_length_distribution}.

\noindent\textbf{\textit{Sample Filtering for Traceability}}.
With each question paired with a \textit{single} context (either generated or retrieved) established in the initial stage, the reader generates the corresponding candidate answer, as shown in Step 2 of \Cref{fig_dataset_construction}.
To unravel the mechanisms of LLMs in context merging, it is essential to ensure the \textit{traceability}, i.e., the output answer is derived from the input context, rather than the intrinsic parametric knowledge of the LLMs.
To achieve this, we only keep samples in which both the generated and retrieved contexts exactly include their respective generated answers, exemplified by $a_{\phi}^{\gamma} {\in} d_1^{\gamma}$, where $\in$ denotes $d_1^{\gamma}$ contains the answer string $a_{\phi}^{\gamma}$.
This practice is grounded in the findings of \citet{chen2022rich,xie2023knowledgeconflicts}, which demonstrate that in the presence of external context, LLMs mostly rely on external context rather than their intrinsic parametric knowledge.
Despite these efforts, the complete elimination of the influence from parametric knowledge remains challenging. 
Therefore, we separately examine its impact in \S\ref{sec 5.1}, where we demonstrate that it has a negligible effect on our conclusions.

\noindent\textbf{\textit{Building Context-Conflicting Dataset}}.
Having obtained answers for each type of context, we are now positioned to construct our \textbf{c}ontext-\textbf{c}onflicting (CC) datasets, as depicted in Step 3 of Figure \ref{fig_dataset_construction}.
Initially, We employ the exact match metric \cite{yu2022genread} to evaluate the correctness of candidate answers derived from contexts, considering an answer correct if its normalized form matches any of the golden answers.

Subsequently, the context-conflicting datasets are composed of samples for which only one of the two types of contexts, either generated or retrieved, yields the correct answer, thereby ensuring the \textit{exclusivity}.
Notably, each dataset comprises two distinct subsets: \textbf{AIG}, consisting of samples with correct \textbf{a}nswers only \textbf{i}n the \textbf{g}enerated context; and \textbf{AIR} comprising samples with correct \textbf{a}nswers only \textbf{i}n the \textbf{r}etrieved context.

\subsection{Statistics of Datasets}
\begin{table}[t]
    \centering
    \begin{adjustbox}{max width=\linewidth}
    \begin{tabular}{lrrrr}
        \toprule
    	\multirowcell{2.5}{Generator\\ \&Reader} & \multicolumn{2}{c}{NQ (12367)} & \multicolumn{2}{c}{TQA (20150)} \\
        \cmidrule(lr){2-3}  \cmidrule(lr){4-5}
        & NQ-AIG & NQ-AIR & TQA-AIG & TQA-AIR \\
    	\midrule
        GPT-4  & 1120 & 763 & 1712 & 681 \\
        GPT-3.5  & 1337 & 857 & 2389 & 1042 \\
        Llama2-13b & 1441 & 1336 & 2982 & 2091 \\
        Llama2-7b & 1423 & 1381 & 3064 & 2604 \\
        \midrule
        Avg. Prop.  & 10.8\% & 8.8\% & 12.6\% & 8.0\% \\
    	\bottomrule
    \end{tabular}
    \end{adjustbox}
    \caption{Dataset size across LLMs. ``Avg. Prop.'' shows average proportions of subsets to original datasets.}
    \label{Statistics}
\end{table}

For each reader-generator pair, we respectively construct context-conflicting datasets from test and dev
sets of NQ and TQA: NQ-CC (NQ-AIG + NQ-AIR) and TQA-CC (TQA-AIG + TQA-AIR).

We initially adopt a typical and simple setting in which an LLM serves as both the generator and reader.
\Cref{Statistics} provides statistics for the constructed subsets corresponding to various LLMs, including GPT-4 (gpt-4-0613), GPT-3.5 (gpt-3.5-turbo-0613), Llama2-7b/13b (Llama2-7b/13b-chat \citep{touvron2023llama}). 
The statistics show that the context-conflicting subsets form a substantial part of the datasets, underscoring the need to investigate how LLMs integrate these distinct contexts. 
Notably, GPT-4 has fewer conflicting instances than other LLMs, because of its higher efficacy in answering questions using either solely retrieved or generated contexts.

\Cref{subsection_Other-Generated Contexts} also explores a more complex scenario in which the generator and reader are distinct LLMs, with the statistics shown in Appendix \ref{appendix_data_quantities}.

\subsection{Evaluation Metric}
\label{evaluation}

Besides datasets, we also develop metrics to study how LLMs merge generated and retrieved contexts in hybrid approaches.
Specifically, the selection of LLMs towards either generated or retrieved context can be measured by the proportion of answers that exactly match the answer produced solely by the corresponding context, denoted as 
\begin{equation*}
    \rho_\text{gen}{=}\text{avg}(\mathtt{em}(a_{\phi}, a_{\phi}^{\varrho})),\quad \rho_\text{ret}=\text{avg}( \mathtt{em}(a_{\phi}, a_{\phi}^{\gamma}))
\end{equation*}
where $\mathtt{em}(a,b)$ returns 1 if $a$ exactly match $b$, and 0 otherwise.
The proportion of instances where $a_{\phi}$ does not match either $a_{\phi}^{\varrho}$ or $a_{\phi}^{\gamma}$ is negligible to the conclusion in this work, as demonstrated in \Cref{others_ratio}.
To facilitate a simple and efficient experiment, we define a synthesized metric as follows:
\begin{equation}
    \mathrm{DiffGR} = \frac{\rho_\text{gen} - \rho_\text{ret}}{\rho_\text{gen}+\rho_\text{ret}}
\end{equation}
The metric $\mathrm{DiffGR}$, ranging from $[-1,1]$, quantifies the extent of LLMs' tendency to rely on generated contexts over retrieved contexts.
Using AIR as an example, where all correct answers come from retrieved contexts, an ideal $\mathrm{DiffGR}$ is $-1$, i.e., LLMs should always rely on retrieved contexts.

\section{How LLMs Merge Contexts?}
\label{Results and Discussing}
This section conducts experiments on the constructed datasets to investigate the merging mechanism of the LLMs in hybrid approaches.
We first consider a typical setting where the generator and reader share a single LLM, to explore how LLMs merge retrieved and \textit{self-generated} contexts (\S\ref{subsection_Self-Generated Contexts}).
Then, we extend our experiments to include more flexible combinations of generator and reader (\S\ref{subsection_Other-Generated Contexts}), and various retrieval models (\S\ref{sec 4.3}), to investigate their effects.

\subsection{LLMs Prefer Self-Generated Contexts} 
\label{subsection_Self-Generated Contexts}

\begin{table}[t]
    \centering
    \begin{adjustbox}{max width=\linewidth} %
    \begin{tabular}{lrrrr}
        \toprule
        \multirow{2}{*}{\parbox{1.5cm}{Generator\\\&Reader}} & \multicolumn{2}{c}{NQ-CC} & \multicolumn{2}{c}{TQA-CC} \\
        \cmidrule(lr){2-3} \cmidrule(lr){4-5}
        & NQ-AIG & NQ-AIR & TQA-AIG & TQA-AIR \\
        \midrule
        GPT-4   & 91.34 & 17.69  & 94.57 & 19.09 \\
        GPT-3.5   & 91.85 & 14.94    & 94.14 & 18.52 \\
        Llama2-13b   & 90.22 & 18.64  & 92.12 & 20.28 \\
        Llama2-7b  & 70.77 & 21.51   & 81.17 & 22.16 \\
        \bottomrule
    \end{tabular}
    \end{adjustbox}
    \caption{The Exact Match (EM) scores (\%) of hybrid approaches on corresponding context-conflicting datasets.}
    \label{table_correct_ratio}
\end{table}

Our preliminary experiments, in which a single LLM serves both as generator and reader, are designed to explore how LLMs integrate information from retrieved and \textit{self-generated} contexts. 
The LLMs under evaluation are tasked with answering questions using both types of contexts on their corresponding CC datasets.
In all experiments, we employ a \textit{randomized} input sequence of contexts to mitigate the influence of order, which is further discussed in Appendix \ref{appendix_order}.

We begin our analysis by examining LLMs' QA performance on CC datasets to reveal how well can LLMs utilize both types of contexts.
Table \ref{table_correct_ratio} presents the Exact Match scores \citep{yu2022genread} across various LLMs.
Surprisingly, LLMs demonstrate significantly low performance ($\leq 22.16\%$) on AIR subsets, despite the fact that the retrieved context alone consistently yields the correct answer on these subsets.
In contrast, LLMs exhibit strong performance on AIG subsets (most near $90\%$).
Overall, all LLMs exhibit a significant performance gap between AIR and AIG datasets, with a pronounced decline in performance when the correct answers come from retrieved contexts.

\begin{figure}[t]
    \centering
    \includegraphics[width=\linewidth]{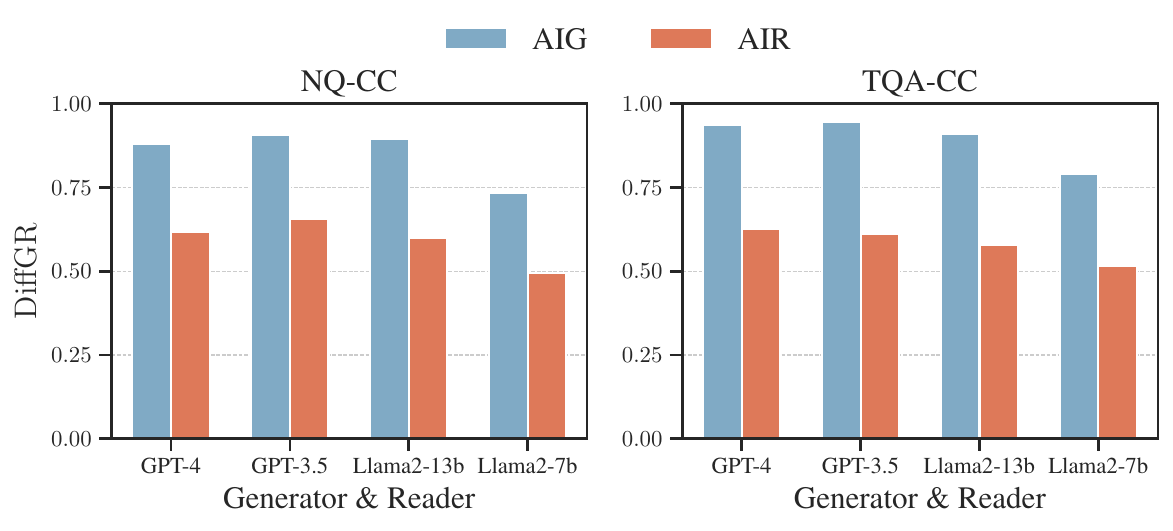}
    \caption{The $\mathrm{DiffGR}$ of LLMs on their corresponding context-conflicting datasets.}
    \label{figure_preference}
\end{figure}

To further reveal LLMs' behavior underlying the QA performance, we trace the source contexts of LLMs' answers using the proposed $\mathrm{DiffGR}$ metric.
An ideal LLM should always rely on retrieved contexts on AIR subsets ($\mathrm{DiffGR=-1}$), and always rely on generated contexts on AIG subsets ($\mathrm{DiffGR=1}$).
Contrary to expectations, Figure \ref{figure_preference} illustrates that LLMs fail to identify the correct information and consistently tend to rely on generated contexts on both AIG and AIR subsets.
This result indicates a pronounced bias in LLMs to \textbf{favor generated contexts, even when they provide incorrect information}.
This bias leads to the insufficient utilization of retrieved contexts mentioned above and highlights a critical challenge for existing LLMs in effectively merging generated and retrieved contexts.
As the bias on AIR subsets has a more direct impact on the performance, the following experiments and analysis will focus on the biases on these subsets to conserve space.
Results on the AIG subsets can be found in Appendix \ref{appendix:AIG}.

\subsection{LLMs Broadly Prefer Generated Contexts} \label{subsection_Other-Generated Contexts}
\begin{figure}[t]
        \centering
            \includegraphics[width=0.9\linewidth]{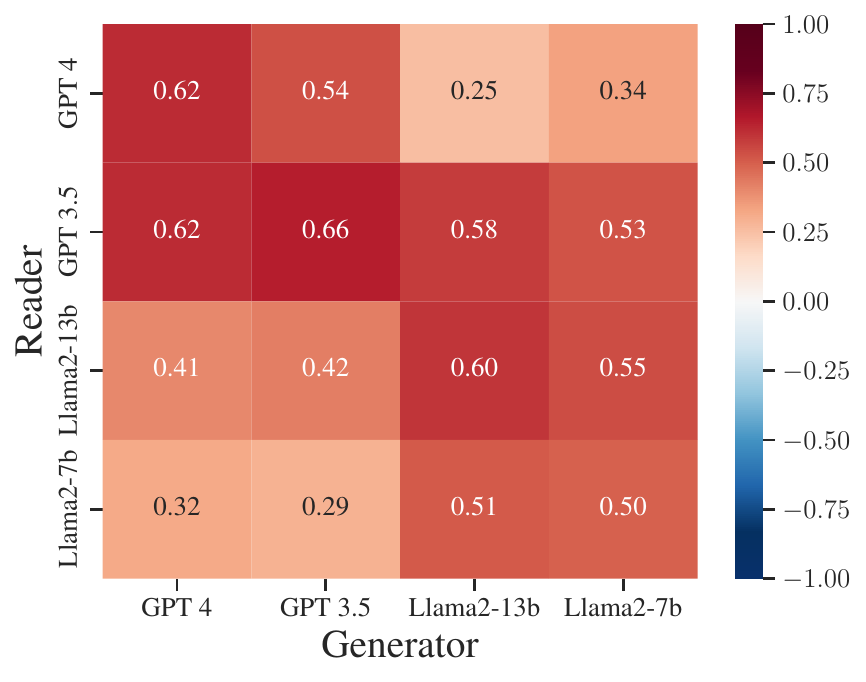}
    \caption{$\mathrm{DiffGR}$ with different (reader, generator) pairs on their corresponding NQ-AIR datasets.}
    \label{preference_reader_generator}
\end{figure}

The above experiments reveal the bias in LLMs to favor the \textit{self-generated} context.
A question emerges: \textit{Do LLMs also prefer contexts generated by other LLMs?}
To investigate this question, this section extends the experiments to more flexible combinations of generators and readers.
This setting is also of practical significance, as recent works have explored the decoupling of generators and readers to achieve modularization of knowledge \citep{luo2023augmented,feng2023cook}.

We construct context-conflicting datasets for each (generator, reader) pair respectively.
The statistics of these datasets are shown in \Cref{appendix_data_quantities}.
Based on these datasets, we then compute $\mathrm{DiffGR}$ metric to examine biases across various (generator, reader) pairs, as shown in Figure \ref{preference_reader_generator}, and observe two notable insights:
\begin{enumerate*}[label=(\roman*)]
    \item \textbf{LLMs are also biased towards contexts generated by other LLMs.} This suggests that such bias in LLMs is widespread and not limited to self-generated contexts.
    \item \textbf{LLMs usually exhibit a stronger bias to contexts generated by themselves.} The sole exception is Llama2-7b, which shows the strongest bias when paired with Llama2-13b as its generator. This phenomenon likely results from their highly similar model structures and training processes \citep{touvron2023llama}.
\end{enumerate*}

\subsection{Consistent Preferences Across Retrievers}
\label{sec 4.3}

The above experiments have demonstrated the bias in LLMs against contexts retrieved by \textbf{Contriever}, a dense retrieval model.
This section further explores whether such bias remains consistent across different retrieval models.
To this end, we incorporate \textbf{BM25} as a representative of sparse retrieval techniques, along with human-labeled golden passages, to mimic an ideal retrieval model (\textbf{Gold}).
For each retrieval model, we respectively construct AIR context-conflicting datasets based on the NQ dev sets from \citet{dpr}, which include the golden passages.
As illustrated in \Cref{table_retrieval}, the results indicate that LLMs consistently display a pronounced bias in favor of generated contexts regardless of the retrieval model used.

\begin{table}[t]
    \centering
    \begin{adjustbox}{max width=\linewidth}
    \small
    \begin{tabular}{lcccc}
        \toprule
        Retriever & Generator \& Reader & $\mathrm{DiffGR}$ \\
        \midrule
        BM25 & Llama2-7b  & 0.5070 \\
        Contriever & Llama2-7b  & 0.5016 \\
        Gold & Llama2-7b  & 0.4656 \\
        \bottomrule
    \end{tabular}
    \end{adjustbox}
    \caption{$\mathrm{DiffGR}$ of different retrievers on respectively AIR datasets constructed from NQ dev set.}
    \label{table_retrieval}
\end{table}

\section{Why LLMs Prefer Generated Contexts}
\label{sec 5}
In this section, we investigate the causes of the observed bias from several perspectives: the effect of parametric knowledge in \S\ref{sec 5.1}, context similarity to the question in \S\ref{section_similarity}, and context completeness in  \S\ref{section_completeness}. 
This section primarily presents the result for Llama2-13b and GPT-3.5, while the results for the other LLMs are provided in \Cref{appendix_Additional_Results}.

\begin{table}[t]
    \centering
    \begin{adjustbox}{max width=\linewidth}
    \begin{tabular}{llrrrr}
        \toprule
    	\multirow{2}{*}{Reader} &  \multirow{2}{*}{Generator} & \multicolumn{2}{c}{NQ (12367)} & \multicolumn{2}{c}{TQA (20150)} \\
        \cmidrule(l){3-4}  \cmidrule(l){5-6}
        & & NQ-AIG & NQ-AIR & TQA-AIG & TQA-AIR \\
    	\midrule
        GPT 3.5 & GPT 3.5 & 500 & 457 & 524 & 322 \\
        GPT 3.5 & Llama2-13b & 271 & 665  & 359 & 574 \\
        Llama2-13b & GPT 3.5 & 1318 & 553 & 1889 & 467 \\
        Llama2-13b & Llama2-13b & 633 & 841 & 928 & 1020 \\
    	\bottomrule
    \end{tabular}
    \end{adjustbox}
    \caption{The number of data in selected subsets where $a_\phi^\text{llm} \neq a_\phi^\gamma \neq a_\phi^\varrho$. More results are shown in \Cref{appendix_tab_filtered_all_llms}.}
    \label{tab_filtered_num}
\end{table}

\subsection{Effect of Parametric Knowledge}
\label{sec 5.1}
Recently, \citet{xie2023knowledgeconflicts} demonstrated that LLMs exhibit a bias towards contexts consistent with their parametric knowledge (or memory), a phenomenon termed \textit{confirmation bias}.
This section explores whether the observed bias arises from the potential consistency between generated contexts and parametric knowledge. 
To investigate this, we select subsets where the answers from ``retrieved context'', ``generated context'', and ``parametric knowledge'' were all different from one another.

To achieve the subset, we first establish the LLM's parametric knowledge about a question using a closed-book QA task, following \citet{xie2023knowledgeconflicts}, which can be expressed as $a_\phi^\text{llm} = \phi(q)$.
Then, we select the cases that satisfy $a_\phi^\text{llm} \neq a_\phi^\gamma \neq a_\phi^\varrho$ from the AIR datasets constructed in \S\ref{Experimental Setup}.
\Cref{tab_filtered_num} shows the number of samples after filtering.

The filtered datasets not only exclude the effect of confirmation bias but also help us identify whether LLMs use parametric knowledge or contexts to answer the question.
\Cref{tab_diffgr_parametric_knowledge} illustrates the proportion of LLMs choosing the answer provided by generated contexts, retrieved contexts, or parametric knowledge.
We observe two key insights: 
\begin{enumerate*}[label=(\roman*)]
    \item The proportion of choosing LLMs' parametric knowledge is very small. This result is consistent with the findings of several previous works \citep{xie2023knowledgeconflicts,chen2022rich}, which found that LLMs mostly rely on the context even when it conflicts with the parametric knowledge.
    \item LLMs still show a significant preference for generated contexts when excluding the influence of parametric knowledge.
    This suggests that the confirmation bias is not the primary cause of the observed bias.
\end{enumerate*}

To ensure rigorous analysis, we use the filtered data in \Cref{tab_filtered_num} for subsequent experiments to eliminate any potential distractions.
\begin{table}[t]
    \centering
    \begin{adjustbox}{max width=\linewidth}
    \begin{tabular}{llccc|cc}
        \toprule
        Reader & Generator & $\rho_\text{gen}$ & $\rho_\text{ret}$ & $\rho_\text{llm}$ & $\mathrm{DiffGR}$ & $\mathrm{DiffGR}$ (ori) \\
        \midrule
        \multicolumn{7}{c}{\textbf{NQ-AIR}} \\
        \cmidrule(l){1-7}
        GPT 3.5 & GPT 3.5 & 67.83 & 12.91 & 0.88 & 0.68 & 0.66 \\
        GPT 3.5 & Llama2-13b & 67.22 & 15.34 & 1.20 & 0.63 & 0.58 \\
        Llama2-13b & GPT 3.5 & 62.39 & 26.76 & 1.08 & 0.40 & 0.42 \\
        Llama2-13b & Llama2-13b & 69.92 & 18.67 & 1.43 & 0.58 & 0.60 \\
        \midrule
        \multicolumn{7}{c}{\textbf{TQA-AIR}} \\
        \cmidrule(l){1-7}
        GPT 3.5 & GPT 3.5 &  72.05 & 16.15 & 1.55 & 0.63 & 0.61 \\
        GPT 3.5 & Llama2-13b & 70.21 & 16.38 & 2.61 & 0.62 & 0.62 \\
        Llama2-13b & GPT 3.5 & 61.88 & 26.34 & 1.71 & 0.40 & 0.35 \\
        Llama2-13b & Llama2-13b & 72.55 & 17.75 & 1.96 & 0.61 & 0.58 \\
        \bottomrule
    \end{tabular}
    \end{adjustbox}
    \caption{The columns $\rho_\text{gen}$, $\rho_\text{ret}$, and $\rho_\text{llm}$ represent the proportion (\%) of responses matching answers from generated contexts, retrieved contexts, and parametric knowledge, respectively. These values are computed on the filtered subset where $a_\phi^\text{llm} \neq a_\phi^\gamma \neq a_\phi^\varrho$, except $\mathrm{DiffGR}$ (ori), which is computed on the original dataset. Results for other LLMs are shown in \Cref{appendix_tab_diffgr_parametric_knowledge}.}
    \label{tab_diffgr_parametric_knowledge}
\end{table}

\subsection{Effect of Text Similarity} \label{section_similarity}
\begin{figure}[t]
    \centering
    \includegraphics[width=\linewidth]{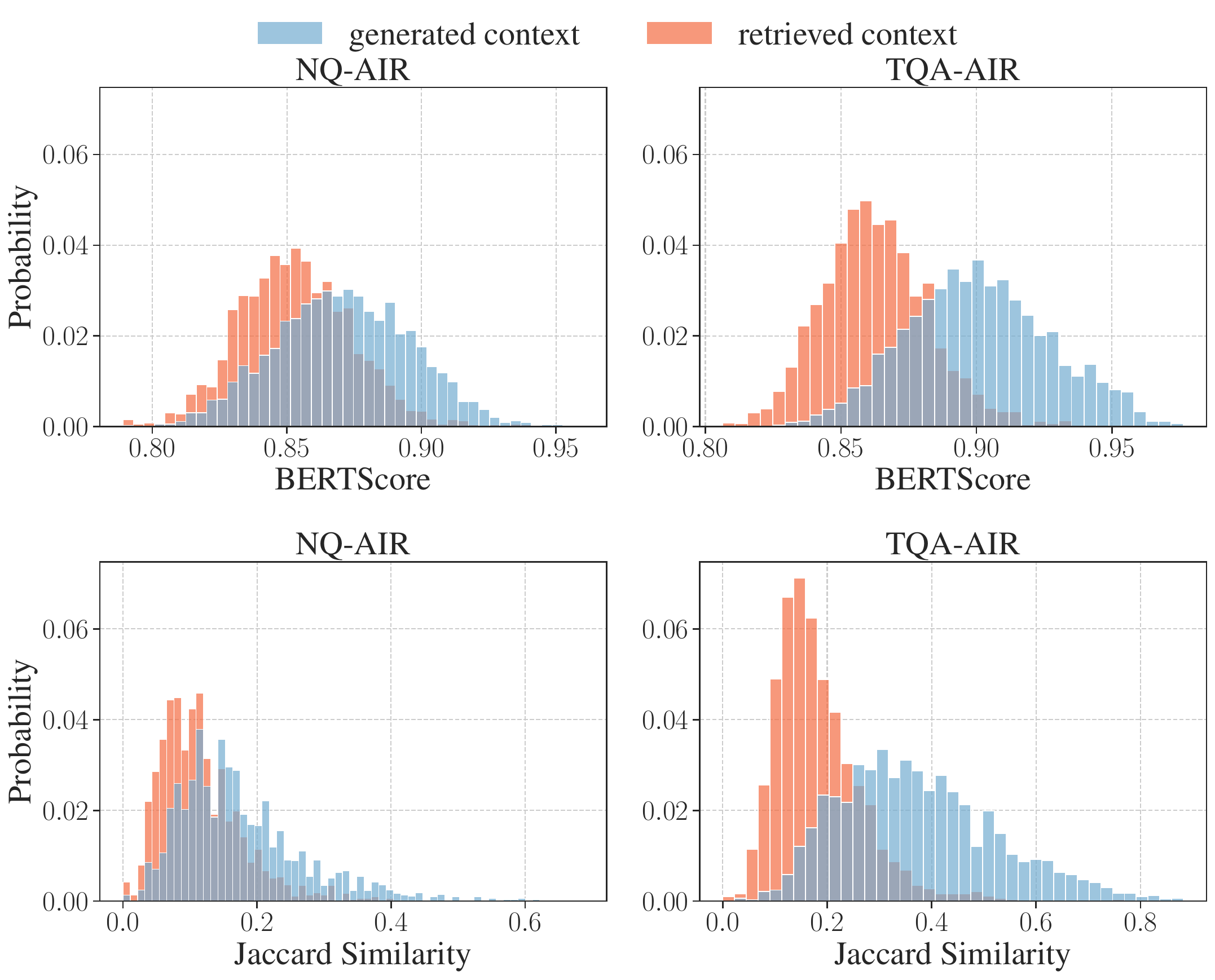}
    \caption{Context-question similarity distribution of generated and retrieved contexts on the union of AIR subsets for different LLMs. More results are shown in \Cref{appendix:Similarity Distribution}.}
    \label{similarity distribution}
\end{figure}

The \textbf{text similarity} between a context and a question can reflect the degree of their relevance.
To investigate the potential effect of the similarity, we employ Jaccard similarity and BERTScore \citep{zhang2019bertscore} to analyze the contexts on the constructed context-conflicting datasets with the reader and generator sharing a single LLM.
\Cref{similarity distribution} shows that generated contexts exhibit a significantly higher similarity to the question on AIR subsets, despite the fact that generated contexts are incorrect on these subsets.
This similarity discrepancy between generated and retrieved contexts persists whether assessed by term-based overlap (average $0.37$ vs. $0.18$ on TQA-AIR) or semantic similarity ($0.90$ vs. $0.86$).

\begin{figure}[t]
    \centering
    \includegraphics[width=\linewidth]{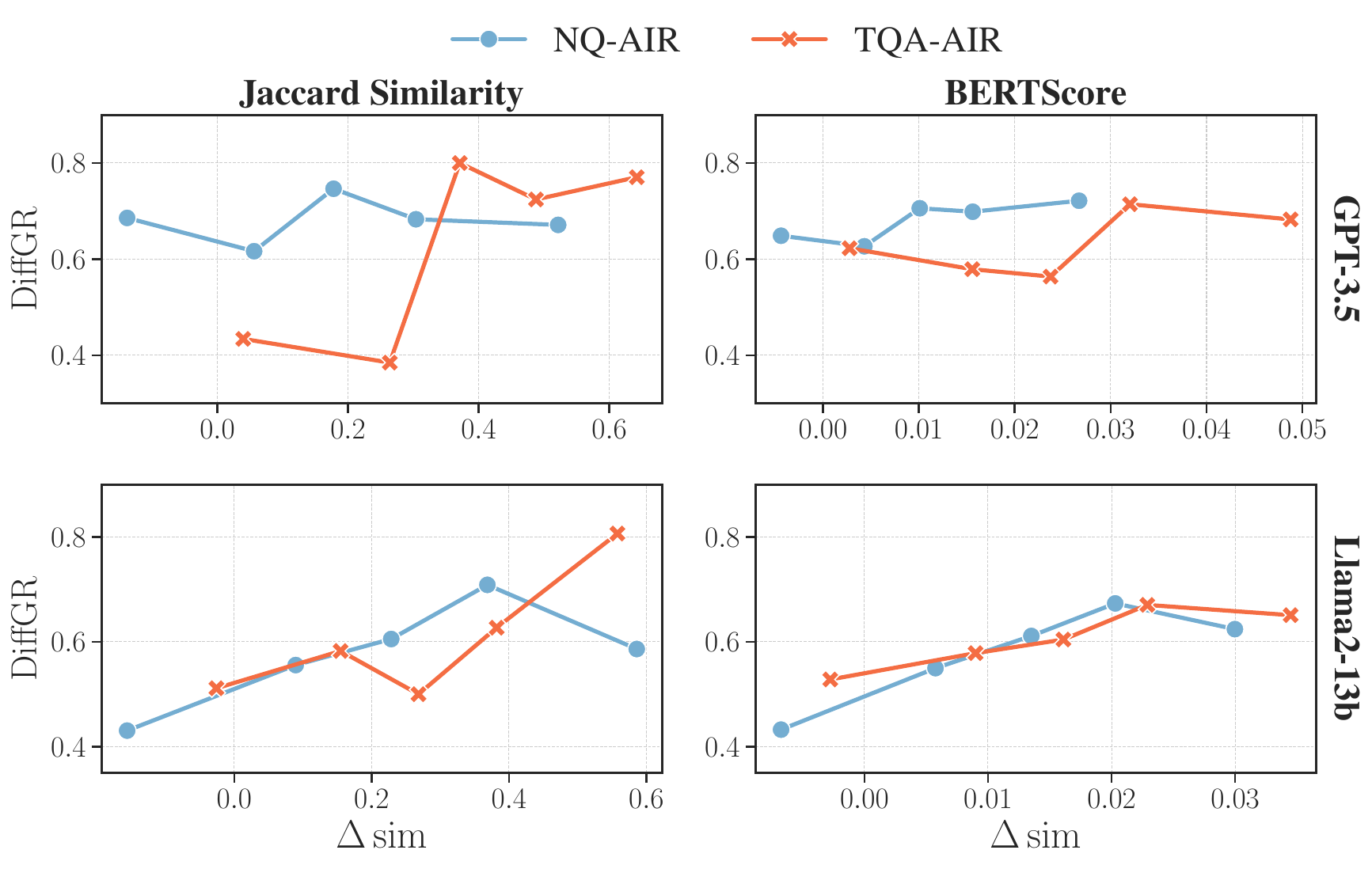}
    \caption{The $\mathrm{DiffGR}$ in slices with different average $\Delta\similarity$. Results for other LLMs are in \Cref{fig:similarity_relation_all_models}.}
    \label{similarity-preference relation}
\end{figure}

To further clarify the influence of the observed similarity discrepancies, we rank the samples according to the similarity gap $\Delta\similarity$ between generated and retrieved contexts. 
$$
\Delta\similarity = \frac{\similarity(q, d^{\varrho})-\similarity(q,d^{\gamma})}{\similarity(q, d^{\varrho})+\similarity(q,d^{\gamma})}
$$
Here, $\similarity(q, d^{\varrho})$ is the similarity between generated context and question, and $\similarity(q,d^{\gamma})$ is for retrieved context.
Then, we divide the dataset into $n$ ($n=5$ here\footnote{Similar results and observations are found with other $n$.}) slices with an equal number of samples. 
Ensuring that each slice contains an equal number of samples helps to avoid fluctuations caused by sample size variations.

Figure \ref{similarity-preference relation} illustrates the relationship between the average $\Delta\similarity$ within each slice and the corresponding $\mathrm{DiffGR}$\footnote{We also tried to manipulate the context similarity to investigate its impact, but we found it challenging to instruct LLMs to generate contexts with low similarity to the question while still providing an answer to it.}. 
From the results, we observe a general trend that \textbf{LLMs exhibit an increased bias to generated contexts on slices with a larger average similarity gap}, which indicates that text similarity is a significant factor in the preference for generated contexts.
These findings suggest that generated contexts should be applied with greater caution to mitigate the influence of highly relevant but misleading information.

To facilitate understanding why the similarity affects LLMs' preference, we include some examples in \Cref{case_similarity}.
From these cases, we observe that contexts with higher similarity often support candidate answers more straightforwardly, for instance, by mirroring the phrasing used in the questions.
Conversely, the contexts with low similarity introduce more challenges, often necessitating an understanding of synonyms and even some inferences.

\subsection{Effect of Context Completeness} 
\label{section_completeness}

In all the above experiments, 
there is a key difference between generated and retrieved contexts that may affect the context preference: \textbf{semantic and} \textbf{sentence completeness}.
Concretely, current retrieval systems typically employ fixed-length truncation to divide a complete article into multiple passages, which serve as the fundamental units for retrieval tasks \citep{dpr,wang2019multi,zhu2021retrieving}.
This truncation often results in retrieved contexts with incomplete semantic meaning, as well as sentences that are cut off at beginnings or endings. 
In contrast, generated contexts in the above experiments are naturally produced by LLMs (\textbf{Nature}), resulting in enhanced semantic and sentence completeness.

\begin{table}[t]
    \centering
    \begin{adjustbox}{max width=\linewidth}
    \begin{tabular}{lccccc}
        \toprule
        \multirow{2.5}{*}{Context} &  \multicolumn{2}{c}{Completeness} &  \multicolumn{2}{c}{Similarity} & \multirow{2.5}{*}{Length} \\
        \cmidrule(lr){2-3} \cmidrule(lr){4-5}
        & Sentence & Semantic & Jaccard & BERTScore &  \\
        \midrule
        Retrieved & \ding{55} & \ding{55}  &  0.1162 & 0.8552 & 107.1 \\
        \midrule
         Nature  & \ding{52} & \ding{52}   & 0.1748 & 0.8730 & 105.7 \\
        S-Trunc. & \ding{52} & \ding{55}  & 0.1733 & 0.8730 &  105.8 \\
        Trunc. & \ding{55} & \ding{55}   & 0.1769 & 0.8736  & 107.1 \\
        \bottomrule
    \end{tabular}
    \end{adjustbox}
    \caption{Average length and similarity of contexts with different completeness on NQ-AIR dataset for GPT-3.5 (more details in \Cref{table_completeness_statistic_full}).}
    \label{table_completeness_statistics}
\end{table}

\begin{table}[t]
    \centering
    \begin{adjustbox}{max width=\linewidth}
    \begin{tabular}{r@{\hskip 4pt}c@{\hskip 4pt}lcccc}
        \toprule
        \multicolumn{3}{c}{\multirow{2.5}{*}{Context Pair}} & \multicolumn{2}{c}{NQ-AIR} & \multicolumn{2}{c}{TQA-AIR}  \\
        \cmidrule(lr){4-5} \cmidrule(lr){6-7}
        & & & GPT-3.5 & Llama2-13b & GPT-3.5 & Llama2-13b  \\
        \midrule
        Nature & vs. & Ret. & \textbf{0.7519} & \textbf{0.6082} & \textbf{0.6353} & \textbf{0.6207} \\
        S-Trunc. & vs. & Ret. & 0.4864 & 0.2551 & 0.5802 & 0.2779 \\
        Trunc. & vs. & Ret. & 0.4792 & 0.2198 & 0.5663 & 0.2787 \\
        \bottomrule
    \end{tabular}
    \end{adjustbox}
    \caption{$\mathrm{DiffGR}$ with different completeness in generated context. ``Nature'', ``Trunc.'' and ``S-Trunc.'' represent three types of generated contexts with different completeness. ``Ret'' means retrieved contexts.}
    \label{table_completeness_preference}
\end{table}

To investigate the potential effects of completeness on the observed bias, we conduct controlled experiments that vary the semantic and sentence completeness of generated contexts\footnote{We also tried to vary the completeness of retrieved contexts but found it challenging to isolate it from confounding factors like length. This aspect is left for future work.} using the following methods:
(a) \textbf{Truncation (Trunc.)} eliminates the length constraints from the generation prompt of \S\ref{Experimental Setup}, allowing LLMs to generate extended contexts. These generated contexts are then truncated to match the length of retrieved contexts, thereby simulating both semantic and sentence incompleteness of retrieved contexts.
(b) \textbf{Sentence Truncation (S-Trunc.)}:  Based on the method (a), we truncate generated contexts only at the end of a sentence to preserve the sentence completeness, while simulating the semantic incompleteness.

To eliminate the interference of similarity factors, we select questions whose three types of generated contexts have nearly equivalent similarity, with BERTScore differences less than 0.05\footnote{We also tried several other thresholds for similarity differences, e.g. 0.005, all yielding same conclusions.}.
Table \ref{table_completeness_statistics} demonstrates that three types of generated contexts also have similar average lengths after filtering.
This means that both the influences of similarity and length are mitigated, thereby highlighting the principal disparities in semantic and sentence completeness.

We evaluate LLMs' preference between generated versus retrieved context, varying the completeness of generated context, following the same pipeline in \S\ref{subsection_Self-Generated Contexts}.
Table~\ref{table_completeness_preference} presents the $\mathrm{DiffGR}$ with different semantic and sentence completeness in generated contexts.
A comparison between ``Trunc.'' and ``S-Trunc.'' reveals that sentence completeness has a very slight impact on LLMs' preference for generated contexts.
In contrast, comparing ``Nature'' and ``S-Trunc.'', we find a significant increase in bias towards generated contexts that are semantically more complete.
These findings indicate that \textbf{LLMs tend to favor contexts with enhanced semantic completeness}, underscoring the necessity to investigate improved passage segmentation methods that maintain semantic completeness for current retrieval-augmented LMs.

\section{Related Work}

\subsection{Generation-Augmented Approaches}
Generation-augmented methods prompt LLMs to generate intermediate contexts for the final response, thereby leveraging their extensive parametric knowledge acquired during the pre-training phase on vast text corpora \citep{knowledge_llm, petroni2019llmknowledge}. 
These generated contexts may encompass various types of knowledge, such as background knowledge \citep{sun2022recitation, yu2022genread}, commonsense knowledge \citep{liu2021generated}, domain-specific knowledge \citep{feng2023cook, luo2023augmented}, and chain-of-thought reasoning processes \citep{wei2022cot,kojima2022large}.
Despite their effectiveness, the LLM-generated knowledge may contain hallucinations \citep{chen-etal-2023-beyond,ji2023hallucinationsurvey} due to LLMs' outdated memory \citep{Editing} and limited memory for long-tail knowledge \citep{kandpal2023longtail}.
This inaccuracy information could potentially mislead current retrieval model \citep{dai2023sourcebias} and open-domain question answering systems \citep{pan2023misinformation}.

\subsection{Retrieval-Augmented Approaches}
Retrieval-augmented methods \citep{guu2020realm,lewis2020retrieval,ram2023context,gao2023retrieval_survey} enhance LLMs by integrating relevant documents from external corpora, addressing limitations like the need for knowledge updates \citep{jang2022temporalwiki} and long-tail knowledge \citep{kandpal2023longtail}.
Early methods \citep{guu2020realm, lewis2020retrieval, izacard2022atlas} focused on joint training of LLMs and retrievers. 
Recent studies \citep{ram2023context,shi2023replug} append relevant documents directly to the input while keeping LLMs static.
Despite their effectiveness, these methods face challenges with irrelevant retrievals and incomplete knowledge coverage \citep{yu2023chainofnote,mallen2022when}.
These noisy retrieval results can misguide LLMs \citep{mallen2022when, yoran2023retrieval_robust, boundary}.

\subsection{Hybrid Approaches and Knowledge Conflicts}
Recent works investigate merging retrieved and generated contexts to leverage both sources of knowledge \citep{abdallah2023generator,yu2022genread, zhang2023merging} and achieve improved performance over those relying solely on a single information source.
However, conflicts between diverse sources can impede the effectiveness of information integration \cite{zhang2023merging}
Current research on knowledge conflicts primarily focuses on two aspects: \textit{context-memory conflict} and \textit{inter-context conflict} \citep{xu2024knowledge}.
Regarding context-memory conflict, \citet{xie2023knowledgeconflicts,chen2022rich} find that LLMs are highly receptive to the input contexts rather than their internal memory.
Concerning inter-context conflict, \citet{chen2022rich} demonstrates that LLMs tend to rely on a few most relevant retrieved contexts.
Additionally, \citet{xie2023knowledgeconflicts} reveals that LLMs favor contexts consistent with their parametric knowledge when confronted with both supporting and opposing contexts.
In contrast to these studies which are limited to conflicts within a single type of context, our work further considers conflicts between generated and retrieved contexts and reveals a bias in LLMs.

\section{Conclusion and Future Work}
In this study, we propose a framework to investigate the underlying mechanisms by which LLMs merge retrieved and generated contexts.
Our results reveal a pronounced bias towards generated contexts in several LLMs (GPT 3.5/4 and Llama2-7b/13b).
We further identify two key factors that may contribute to this bias: higher similarity between generated contexts and questions, and the semantic incompleteness of retrieved contexts. 
Our insights highlight the critical need for advanced integration methods that can validate and leverage information from both sources, moving beyond the current overreliance on generated contexts.

While this study primarily focuses on integrating generated and retrieved contexts, the observed bias also highlights a critical risk of retrieval-augmented LLMs. 
With LLM-generated content increasingly prevalent on the internet, retrieval results may include generated contexts \citep{chen2023combating}.
As this scenario becomes more common, the observed bias of LLMs toward generated contexts implies their susceptibility to misinformation and malicious attacks generated by LLMs. 
This raises concerns about the security of retrieval-augmented systems, a critical problem that is gaining attention in recent works \citep{pan2023misinformation, zou2024poisonedrag}.
Addressing the challenges posed by the widespread presence of generated content, such as the detection of such content, represents a promising direction for future research.

\section*{Limitations}
Our work has the following limitations:
\begin{itemize}[leftmargin=11pt, itemsep=3pt, topsep=0pt, partopsep=0pt, parsep=0pt]
    \item This study is confined to open-domain question answering, a representative knowledge-intensive task. The behavior of LLMs across a broader spectrum of natural language processing tasks remains to be further explored.
    \item This work does not propose specific solutions to effectively mitigate the observed bias, as we focus on revealing the phenomena and analyzing the causes.
    \item To create a controlled environment conducive to analysis, we utilize a single instance for each context type. LLMs face increasingly intricate conflict scenarios when handling multiple contexts from each type. These conflicts emerge not only between retrieved and internally generated contexts but also among the various contexts originating from the same source \citep{chen2022rich,xie2023knowledgeconflicts}.
\end{itemize} 

\section*{Ethics Statement}
\paragraph{Data} All data used in this study are publicly available and do not pose any privacy concerns.

\paragraph{AI Writing Assistance}
In our study, we only employed ChatGPT to polish our textual expressions rather than to generate new ideas or suggestions.

\section*{Acknowledgements}
This work was supported by the National Key R\&D Program of China (2022YFB3103700, 2022YFB3103704), the Strategic Priority Research Program of the Chinese Academy of Sciences (No. XDB0680202), the Innovation Funding of ICT, CAS (E361120), the National Natural Science Foundation of China (No.62172393), and Major Public Welfare Project of Henan Province (No.201300311200).

\bibliography{custom}

\clearpage
\appendix

\section{Detailed Datasets Statistics}

\subsection{Length Control for Generated Contexts}
\label{appendix_length_distribution}
\begin{table}[t]
    \centering
    \begin{adjustbox}{max width=\linewidth}
        \begin{tabular}{lccccc}
            \toprule
            \multirow{2.5}{*}{Dataset} & \multirow{2.5}{*}{Retrieved} & \multicolumn{4}{c}{Generated} \\
            \cmidrule(lr){3-6}
            & & GPT 4 & GPT 3.5 & Llama2-13b & Llama2-7b\\
            \midrule
            NQ & 107.3 & 108.0 & 106.0 & 110.1 & 104.0 \\
            TQA & 106.3 & 107.2 & 104.9 & 105.5 & 102.6 \\
            \bottomrule
        \end{tabular}
    \end{adjustbox}
    \caption{Average lengths of the generated and retrieved contexts. Length is measured in the number of words after punctuation removal.}
    \label{table_length}
\end{table}

In our proposed framework, we regulate the length of generated contexts by incorporating length constraints in the prompt:

\begin{center}
    \resizebox{\linewidth}{!}{
    \begin{tikzpicture}
        \node[text width=\linewidth, align=left, draw, thick, rounded corners=2mm, inner sep=6pt, fill=gray!20] (e1) {
            \textit{Generate a background context from Wikipedia to answer the given question \{\#question\}. Keep the length of the document around $n$ words.}
        };
    \end{tikzpicture}
    \vspace{-9pt}
    }
\end{center}

We observed that GPT 4 effectively controls the output length, whereas other models struggle with this aspect. 
To address this issue in the latter, we employ multiple values of $n$ and select the one that best matches the retrieved context.

\begin{figure*}[!htb]
    \centering
    \includegraphics[width=\textwidth]{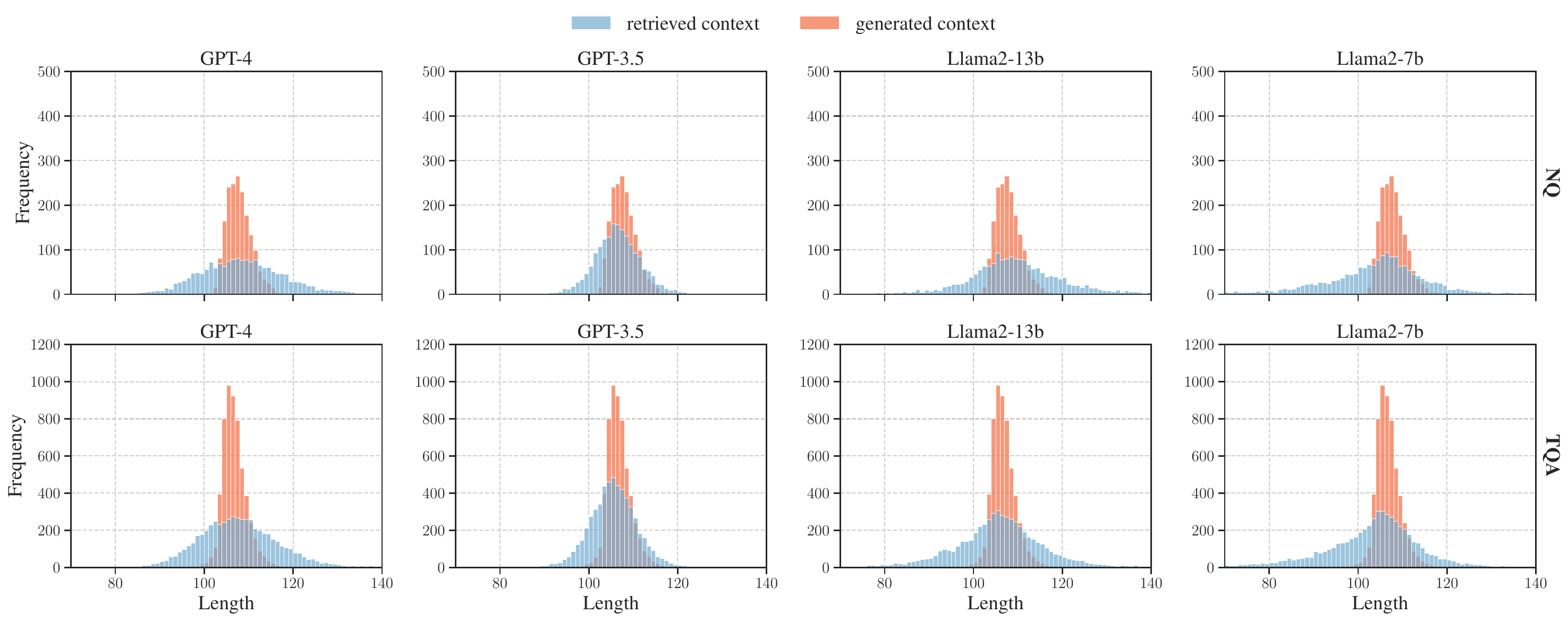}
    \caption{Length distribution of generated and retrieved contexts on different datasets with different generator models.}
    \label{fig:length_distribution_model}
\end{figure*}

As a result, \Cref{table_length} illustrates the average length of different contexts.
\Cref{fig:length_distribution_model} shows the length distribution of retrieved contexts and contexts generated by various LLMs.
The length distribution of retrieved contexts is more concentrated as they consist of text limited to precisely 100 words, along with their titles \citep{dpr}. 
The variation in the length of different retrieved contexts is solely due to the differences in title lengths.

\subsection{Dataset Size} 
\label{appendix_data_quantities}
Table \ref{number_all} presents the data size of context-conflicting datasets corresponding to various generator-reader pairs.
The statistics indicate that conflicting data comprise a substantial proportion across all combinations of generators and readers.
\begin{table}[t]
    \centering
    \begin{adjustbox}{max width=\linewidth}
    \begin{tabular}{llrrrr}
        \toprule
    	\multirow{2}{*}{Reader} &  \multirow{2}{*}{Generator} & \multicolumn{2}{c}{NQ (12367)} & \multicolumn{2}{c}{TQA (20150)} \\
        \cmidrule(l){3-4}  \cmidrule(l){5-6}
        & & NQ-AIG & NQ-AIR & TQA-AIG & TQA-AIR \\
    	\midrule
        GPT 4 & GPT 4 & 1120 & 763 & 1712 & 681 \\
        GPT 4 & GPT 3.5 & 1017 & 922 & - & - \\
        GPT 4 & Llama2-13b & 730 & 1461 & - & - \\
        GPT 4 & Llama2-7b & 600 & 1627 & - & - \\
        GPT 3.5 & GPT 4 & 1514 & 769 & 2701 & 794 \\
        GPT 3.5 & GPT 3.5 & 1337 & 857 & 2389 & 1042 \\
        GPT 3.5 & Llama2-13b & 875 & 1318 & 1781 & 2119 \\
        GPT 3.5 & Llama2-7b & 701 & 1502 & 1471 & 2641 \\
        Llama2-13b & GPT 4 & 2501 & 767 & 4769 & 741 \\
        Llama2-13b & GPT 3.5 & 2211 & 899 & 4210 & 1038 \\
        Llama2-13b & Llama2-13b & 1441 & 1336 & 2982 & 2091 \\
        Llama2-13b & Llama2-7b & 1221 & 1583 & 2567 & 2773 \\
        Llama2-7b & GPT 4 & 2699 & 668 & 5370 & 830 \\
        Llama2-7b & GPT 3.5 & 2435 & 785 & 4813 & 1120 \\
        Llama2-7b & Llama2-13b & 1569 & 1220 & 3526 & 2051 \\
        Llama2-7b & Llama2-7b & 1423 & 1381 & 3064 & 2604 \\
    	\bottomrule
    \end{tabular}
    \end{adjustbox}
    \caption{The data quantities of the constructed subsets for different (Generator, Reader) pairs. NQ and TQA refer to the original datasets (dev+test).}
    \label{number_all}
\end{table}

\section{Additional Results}
\label{appendix_Additional_Results}

\subsection{More Results on AIG Datasets} \label{appendix:AIG}
\begin{figure}[!htb]
    \centering
    \includegraphics[width=0.95\linewidth]{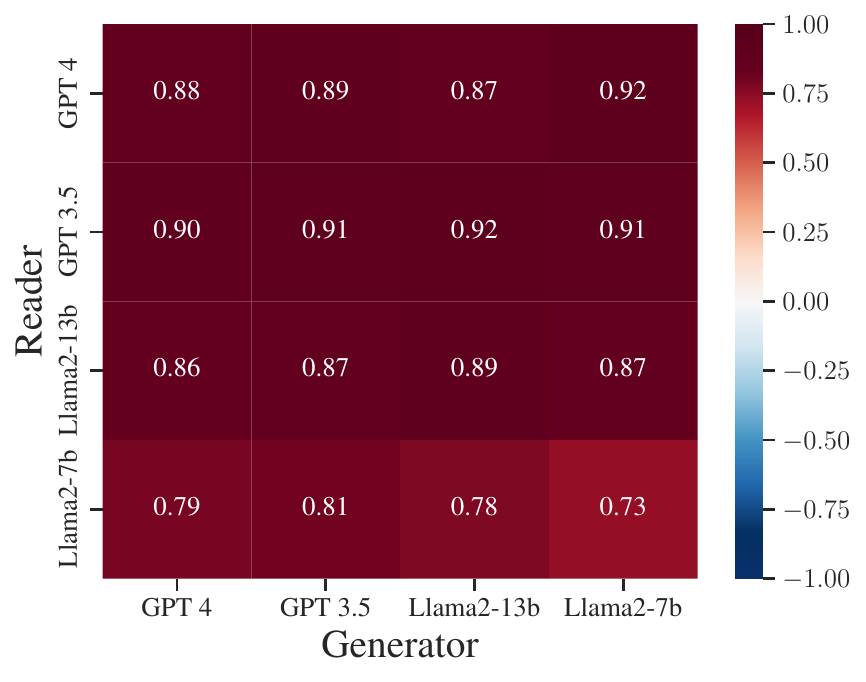}
    \caption{$\mathrm{DiffGR}$ with different (reader, generator) pairs on their corresponding NQ-AIG datasets.}
    \label{fig:diffgr_heatmap_nq_aig}
\end{figure}

Figure \ref{fig:diffgr_heatmap_nq_aig} shows the $\mathrm{DiffGR}$ with different (reader, generator) pairs on their corresponding NQ-AIG datasets.
It can be observed that LLMs show a strong tendency to rely on generated contexts across various (reader, generator) pairs.

\subsection{More Results about Effect of Parametric Knowledge}
\label{appendix_analyse__for_all_LLMs}

\Cref{appendix_tab_filtered_all_llms} illustrates the size of the filtered AIR datasets with the constraint ``$a_\phi^\text{llm} \neq a_\phi^\gamma \neq a_\phi^\varrho$''. 
GPT-4 corresponds to fewer samples because it has better parametric knowledge, i.e., $a_\phi^\text{llm}$ is more likely to be correct.
\Cref{appendix_tab_diffgr_parametric_knowledge} illustrates the proportion of LLMs choosing the answer provided by generated contexts, retrieved contexts, or parametric knowledge.

\begin{table}[t]
    \centering
    \begin{adjustbox}{max width=\linewidth}
    \begin{tabular}{llrrrr}
        \toprule
    	\multirow{2}{*}{Reader} &  \multirow{2}{*}{Generator} & \multicolumn{2}{c}{NQ (12367)} & \multicolumn{2}{c}{TQA (20150)} \\
        \cmidrule(l){3-4}  \cmidrule(l){5-6}
        & & NQ-AIG & NQ-AIR & TQA-AIG & TQA-AIR \\
    	\midrule
        GPT 4 & GPT 4 & 254 & 342 & 259 & 180 \\
        GPT 4 & GPT 3.5 & 220 & 381 & - & - \\
        GPT 4 & Llama2-13b & 160 & 527 & - & - \\
        GPT 4 & Llama2-7b & 124 & 545 & - & - \\ 
        GPT 3.5 & GPT 4 & 653 & 423 & 796 & 237 \\
        GPT 3.5 & GPT 3.5 & 500 & 457 & 524 & 322 \\
        GPT 3.5 & Llama2-13b & 271 & 665  & 359 & 574 \\
        GPT 3.5 & Llama2-7b & 211 & 708 & 279 & 696 \\
        Llama2-13b & GPT 4 & 1563 & 478 & 2321 & 309 \\
        Llama2-13b & GPT 3.5 & 1318 & 553 & 1889 & 467 \\
        Llama2-13b & Llama2-13b & 633 & 841 & 928 & 1020 \\
        Llama2-13b & Llama2-7b & 515 & 1018 & 750 & 1358 \\
        Llama2-7b & GPT 4 & 1896 & 479 & 3163 & 418 \\
        Llama2-7b & GPT 3.5 & 1682 & 557 & 2746 & 591 \\
        Llama2-7b & Llama2-13b & 908 & 884 & 1681 & 1238 \\
        Llama2-7b & Llama2-7b & 753 & 1002 & 1272 & 1538 \\
    	\bottomrule
    \end{tabular}
    \end{adjustbox}
    \caption{The number of data in selected subsets where $a_\phi^\text{llm} \neq a_\phi^\gamma \neq a_\phi^\varrho$.}
    \label{appendix_tab_filtered_all_llms}
\end{table}

\begin{table*}[t]
    \centering
    \begin{adjustbox}{max width=\linewidth}
    \begin{tabular}{llcccccccc}
        \toprule
        \multirow{2}{*}{Reader} & \multirow{2}{*}{Generator} & \multicolumn{4}{c}{NQ-AIR} & \multicolumn{4}{c}{TQA-AIR} \\
        \cmidrule(lr){3-6} \cmidrule(lr){7-10}
        & & $\rho_\text{gen}$ & $\rho_\text{ret}$ & $\rho_\text{llm}$ & $\mathrm{DiffGR}$ & $\rho_\text{gen}$ & $\rho_\text{ret}$ & $\rho_\text{llm}$ & $\mathrm{DiffGR}$ \\
        \midrule
        GPT 4 & GPT 4 & 66.08 & 18.71 & 2.05 & 0.5586 & 76.67 & 13.89 & 2.78 & 0.6933 \\
        GPT 4 & GPT 3.5 & 66.40 & 23.62 & 1.84 & 0.4752 & - & - & - & - \\
        GPT 4 & Llama2-13b & 58.63 & 28.84 & 2.85 & 0.3406 & - & - & - & - \\
        GPT 4 & Llama2-7b & 62.20 & 26.42 & 2.39 & 0.4037 & - & - & - & - \\
        GPT 3.5 & GPT 4 & 65.25 & 16.08 & 0.95 & 0.6047 & 68.78 & 17.72 & 3.38 & 0.5902 \\
        GPT 3.5 & GPT 3.5 & 67.83 & 12.91 & 0.88 & 0.6802 & 72.05 & 16.15 & 1.55 & 0.6338 \\
        GPT 3.5 & Llama2-13b & 67.22 & 15.34 & 1.20 & 0.6284 & 70.21 & 16.38 & 2.61 & 0.6217 \\
        GPT 3.5 & Llama2-7b & 64.55 & 16.24 & 1.13 & 0.5979 & 74.14 & 13.65 & 1.87 & 0.6890 \\
        Llama2-13b & GPT 4 & 61.72 & 25.73 & 2.51 & 0.4115 & 57.28 & 33.01 & 1.29 & 0.2688 \\
        Llama2-13b & GPT 3.5 & 62.39 & 26.76 & 1.08 & 0.3996 & 61.88 & 26.34 & 1.71 & 0.4029 \\
        Llama2-13b & Llama2-13b & 69.92 & 18.67 & 1.43 & 0.5785 & 72.55 & 17.75 & 1.96 & 0.6069 \\
        Llama2-13b & Llama2-7b & 69.16 & 18.96 & 1.87 & 0.5697 & 75.04 & 15.54 & 1.62 & 0.6569 \\
        Llama2-7b & GPT 4 & 54.28 & 24.43 & 1.46 & 0.3793 & 47.13 & 33.97 & 3.59 & 0.1622 \\
        Llama2-7b & GPT 3.5 & 52.24 & 27.29 & 1.80 & 0.3138 & 54.31 & 28.76 & 2.20 & 0.3075 \\
        Llama2-7b & Llama2-13b & 61.65 & 19.46 & 3.17 & 0.5202 & 64.46 & 20.76 & 1.86 & 0.5128 \\
        Llama2-7b & Llama2-7b & 60.38 & 21.16 & 2.20 & 0.4810 & 66.25 & 18.40 & 3.12 & 0.5653 \\
        
        \bottomrule
    \end{tabular}
    \end{adjustbox}
    \caption{The columns labeled $\rho_\text{gen}$, $\rho_\text{ret}$, and $\rho_\text{llm}$ respectively represent the proportion (\%) of responses that match the answer provided by generated contexts, retrieved contexts, and internal parametric knowledge of LLMs.}
    \label{appendix_tab_diffgr_parametric_knowledge}
\end{table*}

\subsection{Effect of Context Order} \label{appendix_order}
\begin{table}[t]
    \centering
    \begin{tabular}{lcc}
        \toprule
        Order & NQ-AIR & TQA-AIR \\
        \midrule
        generated-first & 0.699 & 0.682 \\
        retrieved-first & 0.665 & 0.556 \\
        random & 0.691 & 0.586 \\
        \bottomrule
    \end{tabular}
    \caption{$\mathrm{DiffGR}$ with different context order on NQ-AIR and TQA-AIR datasets. GPT-3.5 serves as both the generator and reader.}
    \label{order}
\end{table}
In the above experiments, retrieved and generated contexts are presented in random order.
Previous studies \cite{xie2023knowledgeconflicts,liu2023lost,lu2021fantastically} have found that the model may be sensitive to the order of the input contexts.
In their experiments, the input context was either all retrieved  \cite{liu2023lost} or all generated \cite{xie2023knowledgeconflicts}.
We conducted experiments to investigate whether the context order impacts the preference for the generated context.
The generated and retrieved contexts are concatenated with three different orders: generated-first, retrieved-first, and random.
To control the cost of API, this section conducts experiments on the context-conflicting datasets from only the test sets of NQ and TQA.
We compute the $\mathrm{DiffGR}$ with different context orders respectively.

As shown in Table \ref{order}, across all context orders, LLMs consistently show a strong tendency to favor generated contexts.
When the retrieved context is positioned first, there is a slight reduction in $\mathrm{DiffGR}$.
This reduction may result from the LLMs' preference for generated contexts being partially offset by their bias towards the top context \citep{liu2023lost, xie2023knowledgeconflicts}.

\begin{table}[thb]
    \centering
    \begin{adjustbox}{max width=\linewidth}
    \small
    \begin{tabular}{lcc}
        \toprule
        & NQ-AIR & TQA-AIR \\
        \midrule
        Llama2-13b & 632 & 702 \\
        GPT-3.5 &  323 & 181 \\
        \bottomrule
    \end{tabular}
    \end{adjustbox}
    \caption{The number of samples filtered with a BERTScore difference of less than 0.05.}
    \label{table_completeness_filtered_size}
\end{table}

\subsection{More Results about Effect of Context Completeness}
\begin{figure*}[!htb]
    \centering
    \includegraphics[width=\textwidth]{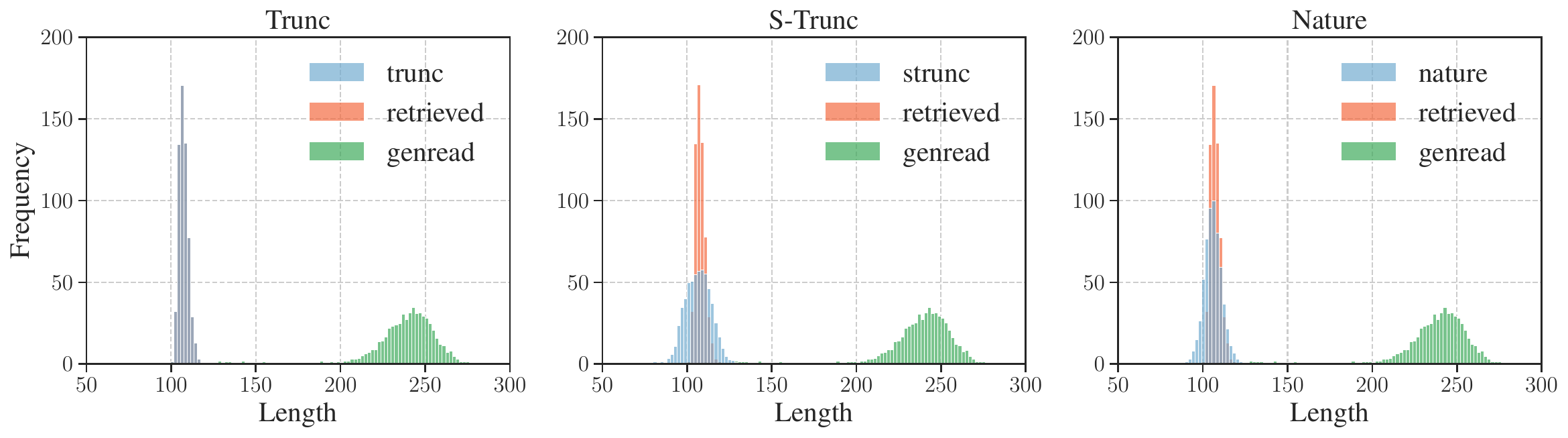}
    \caption{Length distribution of generated and retrieved contexts on the NQ dataset with GPT-3.5 as the generator. ``genread'' represents the contexts generated by the original GenRead method \citep{yu2022genread}. ``trunc'', ``strunc'', and ``nature'' are the generated contexts using three different methods to control the length.}
    \label{fig:length_distribution_control_methods}
\end{figure*}

\textbf{(a) Length Distribution.} In \S\ref{section_completeness}, we employ three methods, ``Nature'', ``Trunc.'' and ``S-Trunc.'', to vary the completeness of generated contexts, while controlling the length at the same time.
Figure \ref{fig:length_distribution_control_methods} illustrates the length distribution for generated contexts corresponding to these methods. 
From the results, we can observe that the contexts generated by original GenRead \citep{yu2022genread} are significantly longer compared to the retrieved contexts. 

\noindent\textbf{(b) Similarity and Length Control.} \Cref{table_completeness_filtered_size} presents the data quantity after filtering out samples where three types of generated contexts exhibit significant differences in similarity.
\Cref{table_completeness_statistic_full} illustrates the average similarity and completeness of three types of generated contexts. ``Nature'', ``Trunc.'' and ``S-Trunc.'' result in contexts with Comparable average length and similarity, with preliminary differences in completeness.

\begin{table*}[t]
    \centering
    \begin{adjustbox}{max width=\linewidth}
    \begin{tabular}{llcccccccc}
        \toprule
        \multicolumn{2}{c}{\multirow{2}{*}{Context}} & \multicolumn{2}{c}{Completeness} & \multicolumn{3}{c}{NQ-AIR} & \multicolumn{3}{c}{TQA-AIR} \\
        \cmidrule(lr){3-4} \cmidrule(lr){5-7} \cmidrule(lr){8-10}
        & & Sentence & Semantic & Length & Jaccard & BERTScore & Length & Jaccard & BERTScore \\
        \midrule
        \multirow{4}{*}{Llama2-13b} &  
        Retrieved & \ding{55} & \ding{55} & 107.8 & 0.1157 & 0.8534 & 106.6 & 0.1832 & 0.8630 \\
        & Nature & \ding{52} & \ding{52} &  109.4 & 0.1859 & 0.8755 & 105.0 & 0.3178 & 0.8909 \\
        & S-Trunc &  \ding{52} & \ding{55} & 106.1 & 0.1734 & 0.8719 & 105.0 & 0.2868 & 0.8833 \\
        & Trunc & \ding{55} & \ding{55} &  107.8 & 0.1765 & 0.8726 & 106.6 & 0.2890 & 0.8837 \\
        \midrule
        \multirow{4}{*}{GPT-3.5} & 
        Retrieved & \ding{55} & \ding{55} &  107.1 & 0.1162 & 0.8552 & 106.3 & 0.1836 & 0.8622 \\
        & Nature  & \ding{52} & \ding{52} & 105.7 & 0.1748 & 0.8730 & 105.0 & 0.3993 & 0.9044 \\
        & S-Trunc & \ding{52} & \ding{55} & 105.8 & 0.1733 & 0.8730 & 104.8 & 0.4024 & 0.9044 \\
        & Trunc & \ding{55} & \ding{55} & 107.1 & 0.1769 & 0.8736 & 106.3 & 0.4027 & 0.9044 \\
        \bottomrule
    \end{tabular}
    \end{adjustbox}
    \caption{Average length and similarity of contexts for GPT-3.5 and Llama2-13b models. The three types of generated contexts exhibit similar average lengths and similarity, with the primary distinction being their completeness.}
    \label{table_completeness_statistic_full}
\end{table*}

\subsubsection{Cases about Completeness}
\Cref{case_completeness} provides some examples to facilitate the understanding of completeness.
From the cases, we observe that retrieved contexts and ``Trunc.'' often contain incomplete sentences.
Additionally, compared to ``S-Trunc'', ``Nature'' typically exhibits greater semantic completeness. Specifically, ``Nature'' often encompasses a full logical structure of an article, including an introduction, discussion, and conclusion, whereas ``S-Trunc'' may terminate abruptly.

\begin{table*}[t] 
    \centering
    \begin{adjustbox}{max width=\linewidth} 
    \begin{tabular}{c}
        \toprule
        Example \\
        \midrule
        \parbox{16cm}{
            \textbf{Question:} In World War I, what did the Allies call the defence line that was known as the 'Siegfried Line' to the Germans? \\
\textbf{Answer:} ['Hindenberg Line', 'Hindenburg line', 'Battles of the Hindenburg Line', "'HINDENBURG LINE'", 'Advance to the Hindenburg Line', 'Hindenburg Line'] \\
\textbf{Retrieved context:} Title: Siegfried Line Content: Siegfried Line The term Siegfried Line refers to two different German defensive lines, one during the First World War and the other during the Second World War. The original was a line of forts and tank defences built by Germany in northern France during 1916–17, known in German as the Siegfriedstellung ("Siegfried Position"). The British referred to it as the "Hindenburg Line", a name subsequently adopted by the rest of the Allies. After WW1 the German Fortifications East of the Rhein were required to be demolished due to the Treaty of Versailles. The second, commonly known in German as the \\
\textbf{Generated context (Nature):} The Allies referred to the German defence line known as the 'Siegfried Line' as the 'Westwall' during World War I. Constructed between 1936 and 1940, the Westwall was a system of fortifications built along Germany's western border. It stretched from the Swiss border to the Dutch border, covering approximately 630 kilometers. The line consisted of concrete bunkers, tank traps, and anti-tank ditches, aiming to deter any potential invasion from the west. The Allies recognized the formidable nature of this defensive barrier and thus referred to it as the Westwall, while the Germans named it the Siegfried Line, after the mythical hero Siegfried from German folklore. \\
\textbf{Generated context (Trunc.):} The Allies referred to the German defence line known as the "Siegfried Line" as the "Westwall" during World War I. The Westwall was a series of fortifications constructed by Germany along its western border, stretching from the Swiss border to the Netherlands. It was named after the German Emperor, Wilhelm II, who was commonly referred to as "Kaiser Wilhelm" or "Siegfried" by the Allies.
The construction of the Westwall began in the 1930s as a response to the growing tensions in Europe and the threat of war. The line consisted of a network of bunkers, pillboxes, tank traps, and other defensive structures, strategically positioned \\
\textbf{Generated context (S-Trunc.):} The Allies referred to the German defence line known as the "Siegfried Line" as the "Westwall" during World War I. The Westwall was a series of fortifications constructed by Germany along its western border, stretching from the Swiss border to the Netherlands. It was named after the German Emperor, Wilhelm II, who was commonly referred to as "Kaiser Wilhelm" or "Siegfried" by the Allies.
The construction of the Westwall began in the 1930s as a response to the growing tensions in Europe and the threat of war. The line consisted of a network of bunkers, pillboxes, tank traps, and other defensive structures, strategically positioned to impede any potential invasion from the west. \\
        } \\
        \bottomrule
    \end{tabular}
    \end{adjustbox}
    \caption{Examples with retrieved contexts and generated contexts. ``Nature'', ``Trunc.'' and ``S-Trunc.'' represent three types of generated contexts with different completeness. Retrieved contexts often contain incomplete sentences.}
    \label{case_completeness}
\end{table*}

\subsection{More Results about Effect of Text Similarity} \label{appendix:Similarity Distribution}

\textbf{(a) Similarity Metric.}
We employ Jaccard similarity to assess the term-based overlap,  and BERTScore \citep{zhang2019bertscore} for evaluating the semantic similarity between contexts and questions. 
To mitigate the effect of length discrepancies between contexts and questions, we calculate the similarity at the sentence level and then aggregate them to derive the overall context-question similarity.
In this work, we adopt a maximum aggregation strategy due to the single-hop nature of the NQ and TQA datasets, where the majority of questions can be answered using a small subset of sentences.
We also try the average aggregation strategy and observe similar results.

\begin{figure*}[!htb]
    \centering
    \begin{adjustbox}{max width=\linewidth}
    \includegraphics[width=0.95\textwidth]{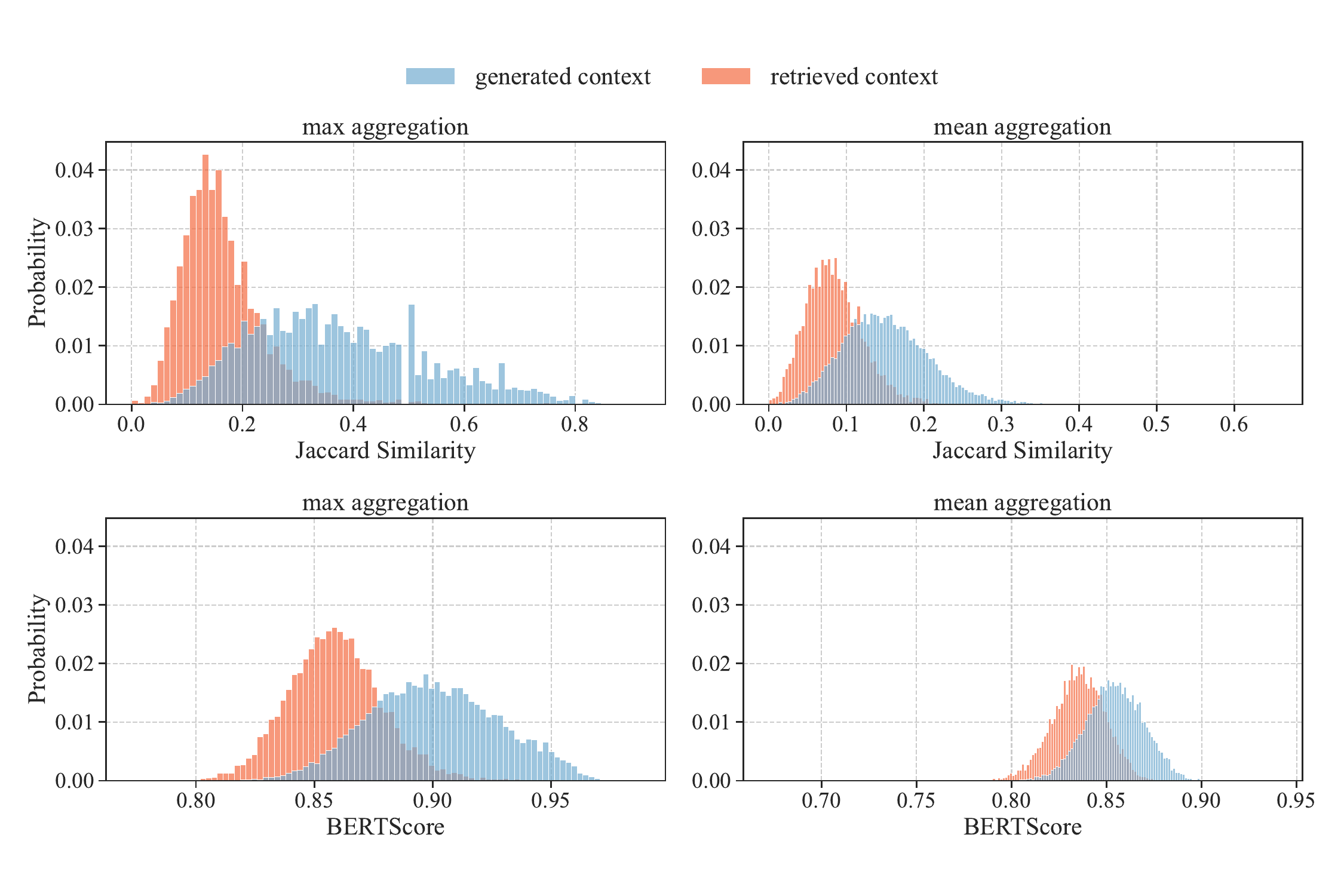}
    \end{adjustbox}
    \caption{Similarity distribution with maximum or mean aggregation strategies. Generated contexts consistently exhibit higher similarity across two aggregation strategies.}
    \label{fig:min_max}
\end{figure*}

Figure \ref{fig:min_max} illustrates the distribution of similarity when employing maximum and average aggregation methods. 
It is observable that the generated contexts exhibit a markedly higher degree of similarity regardless of the aggregation method used. 
Furthermore, this disparity in similarity is more pronounced with maximum aggregation, as contexts typically contain sentences that are irrelevant, which dilute the similarity scores when an average aggregation is applied.

\begin{figure*}[!htb]
    \centering
    \includegraphics[width=0.95\textwidth]{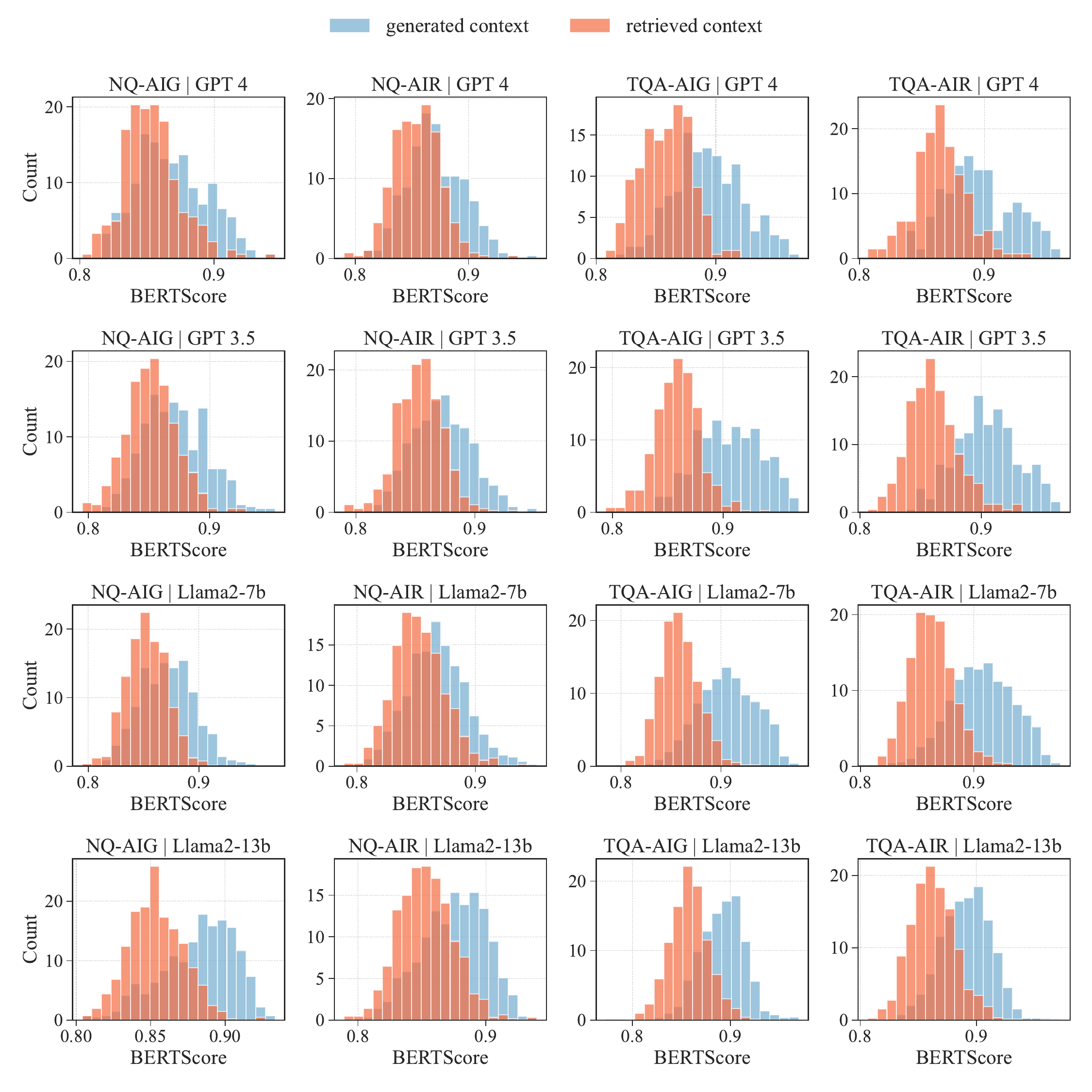}
    \caption{BERTScore distribution of retrieved contexts and contexts generated by different LLMs. All LLM-generated contexts exhibit a higher similarity over retrieved contexts.}
    \label{fig:bertscore_all_models}
\end{figure*}

\begin{figure*}[!htb]
    \centering
    \includegraphics[width=0.95\textwidth]{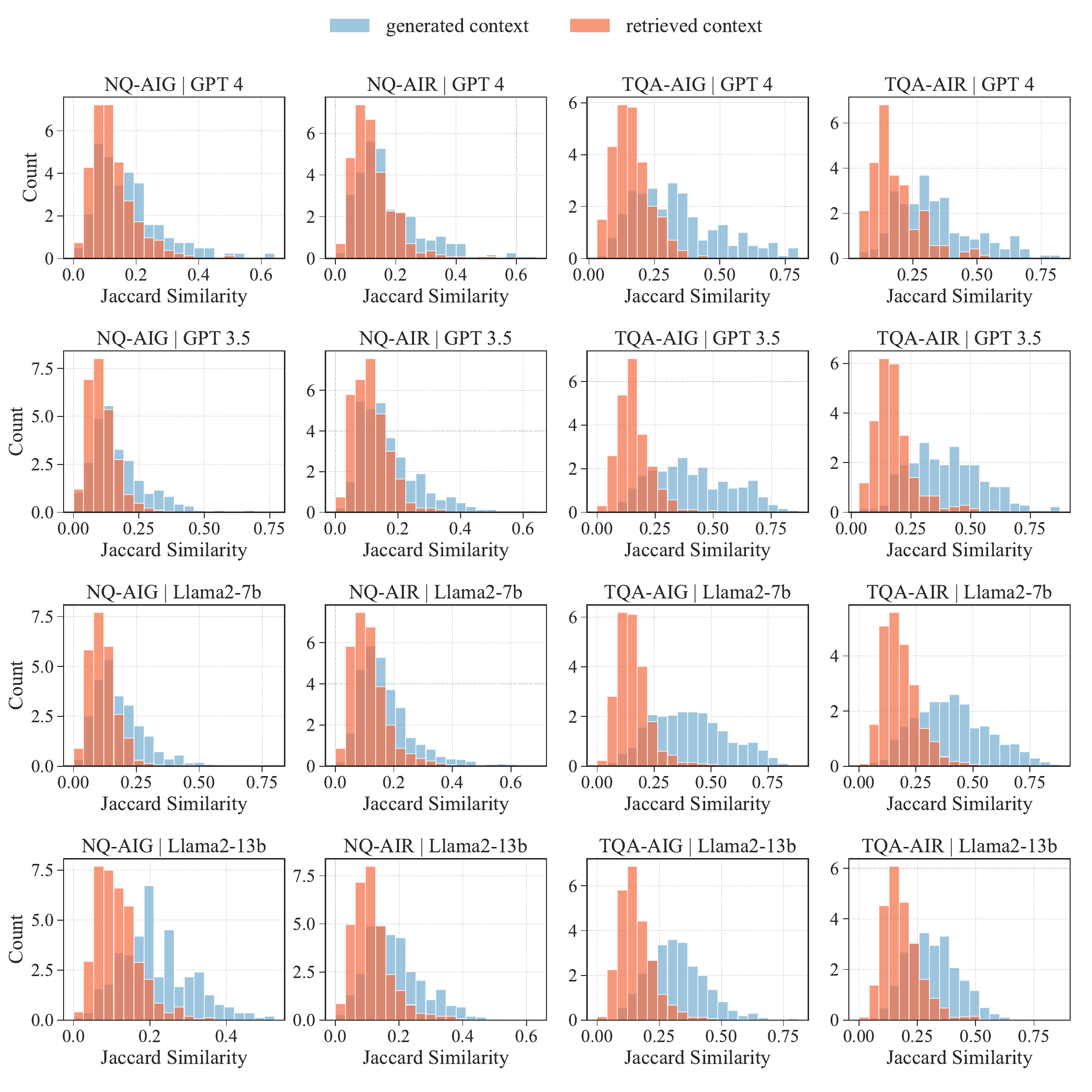}
    \caption{Jaccard Similarity distribution of retrieved contexts and contexts generated by different LLMs. All LLM-generated contexts exhibit a higher similarity over retrieved contexts.}
    \label{fig:jaccard_all_models}
\end{figure*}

\noindent \textbf{(b) Similarity Distribution.} Figure \ref{fig:bertscore_all_models} and \ref{fig:jaccard_all_models} show the similarity distribution of retrieved and generated contexts across various generators.
All LLM-generated contexts exhibit a higher similarity over retrieved contexts.

\begin{figure*}[!htb]
    \centering
    \begin{minipage}{0.49\textwidth}
        \includegraphics[width=\textwidth]{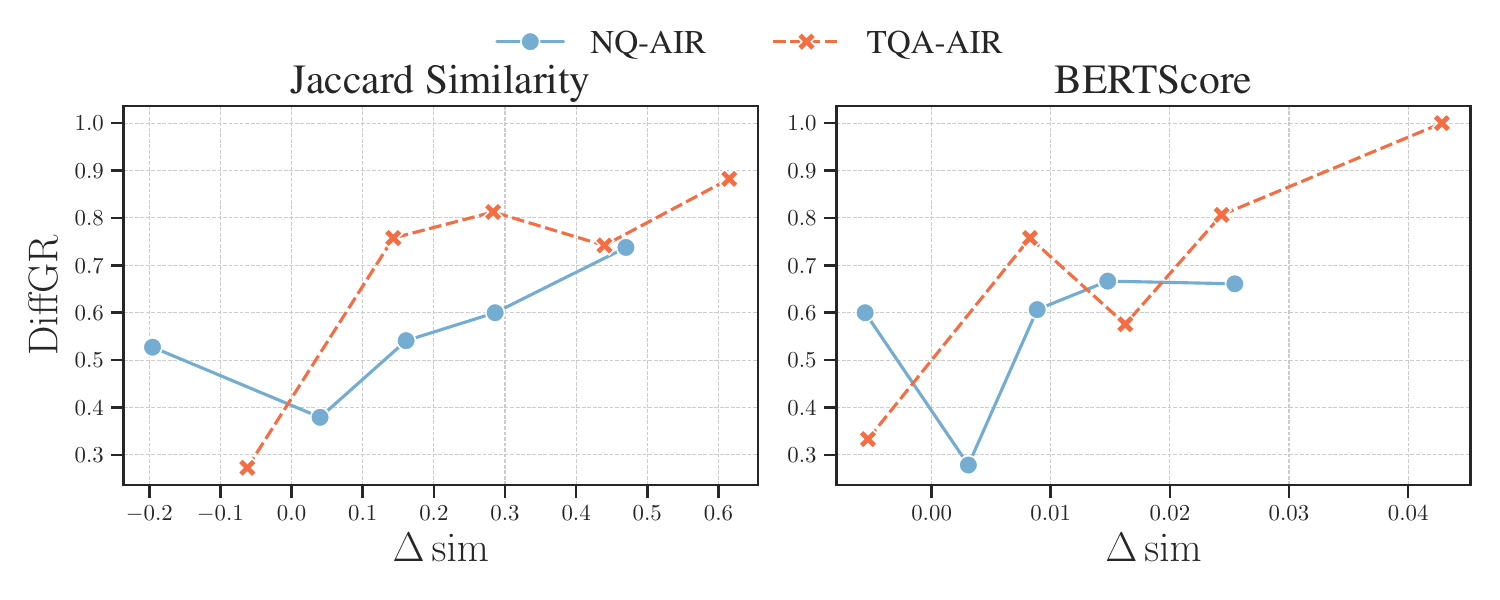}
        \subcaption{GPT 4}
    \end{minipage}
    \begin{minipage}{0.49\textwidth}
        \includegraphics[width=\textwidth]{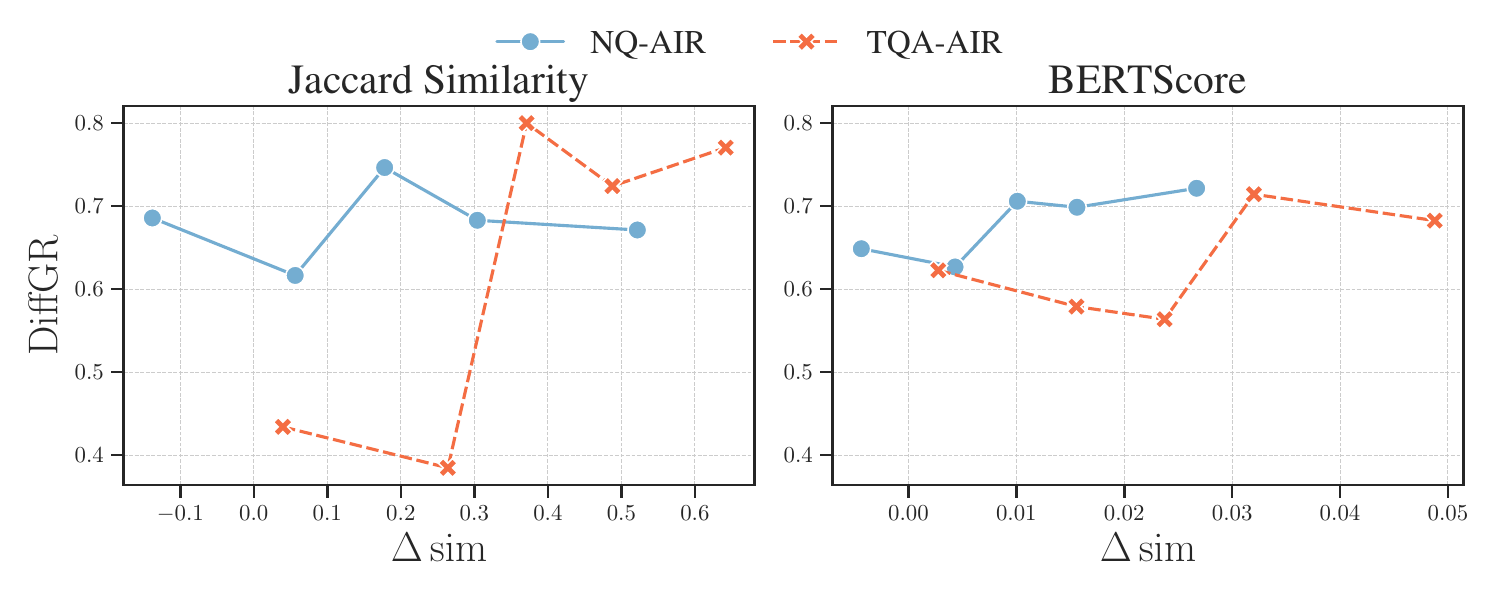}
        \subcaption{GPT 3.5}
    \end{minipage} \\
    \begin{minipage}{0.49\textwidth}
        \includegraphics[width=\textwidth]{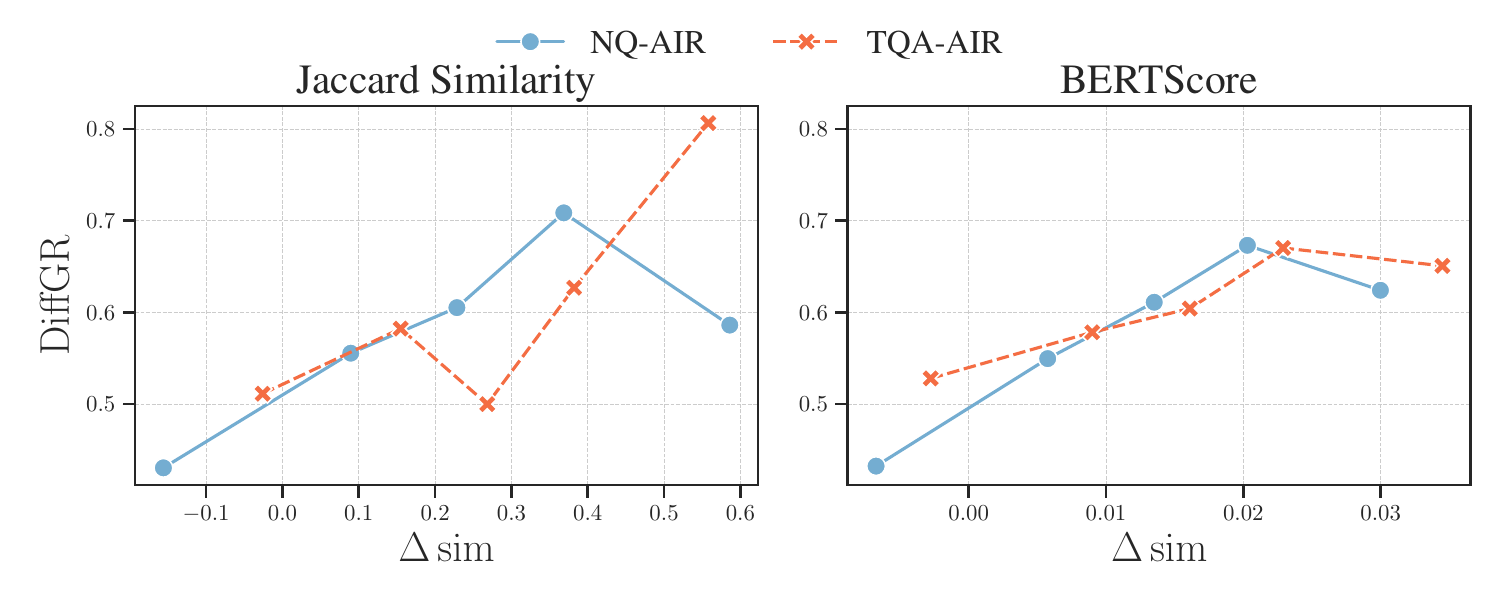}
        \subcaption{Llama2-13b}
    \end{minipage}
    \begin{minipage}{0.49\textwidth}
        \includegraphics[width=\textwidth]{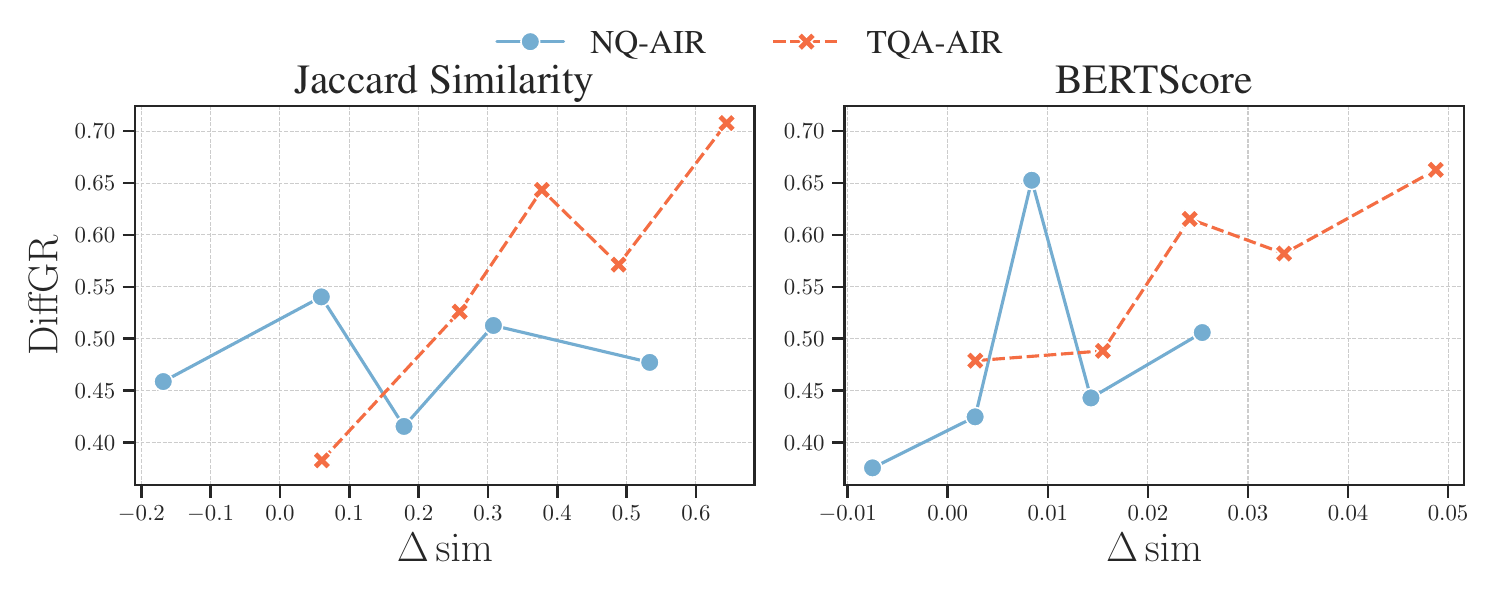}
        \subcaption{Llama2-7b}
    \end{minipage}
    \caption{The generation ratio in slices with different average $\Delta \operatorname{sim}$. $\Delta \operatorname{sim}$ is the difference in similarity between the generated context and the retrieved context. The LLM marked in the figure serves both as a generator and a reader.}
    \label{fig:similarity_relation_all_models}
\end{figure*}

\noindent \textbf{(c) Effect of Similarity.} Figure \ref{fig:similarity_relation_all_models} demonstrates a general trend that on slices with a smaller average similarity gap, LLMs exhibit a reduced preference for generated context.

\noindent \textbf{(d) Cases about Similarity.} Table \ref{case_similarity} shows examples that contain contexts with different similarities to the question.
The contexts with high similarity typically directly support answering by repeating the phrasing in the question.
Conversely, the contexts with low similarity introduce more challenges, often necessitating an understanding of synonyms and even some inferences.
These observations indicate that text similarity can partly reflect the relevance between a question and a context, as well as the difficulty the LLM encounters in identifying potential answers.
\begin{table*}[t] 
    \centering
    \begin{adjustbox}{max width=\linewidth} 
    \begin{tabular}{cll}
        \toprule
        & TQA-AIR Example & TQA-AIG Example \\
        \midrule
        \multirow{1}{*}{Question} & \parbox[c]{6cm}{Between 1959 and 1967 which city was the capital of Pakistan (Islamabad was being built)?} & \parbox[c]{6cm}{Who is the most successful UK solo artist in the USA?} \\
        \midrule
        Golden Answer & Rawalpindi  & Elton John \\
        \midrule
         \multirow{3}{*}{Generated Context} &  \parbox[c]{6cm}{Between 1959 and 1967, the capital of Pakistan was \sethlcolor{pink}\hl{Karachi}. Karachi is the largest city in Pakistan and is located on the southern coast of the country\dots}
         &  
         \parbox[c]{6cm}{\sethlcolor{green}\hl{Elton John} is the most successful UK solo artist in the USA. Born Reginald Kenneth Dwight in 1947, he adopted the stage name Elton John in the late 1960s\dots} \\ 
         & \textbf{Jaccard Similarity}: 0.47 & \textbf{Jaccard Similarity}: 0.69 \\
         & \textbf{BertScore}: 0.93  & \textbf{BertScore}: 0.93  \\
        \midrule
        \multirow{3}{*}{Retrieved Context} &  \parbox[c]{6cm}{\dots was first shifted temporarily to \sethlcolor{green}\hl{Rawalpindi} in the early 60s, and then to Islamabad when essential development work was completed in 1966\dots}
        &  
        \parbox[c]{6cm}{\dots In 2009, \sethlcolor{pink}\hl{Jay Sean}'s single "Down" reached the number one spot on the "Billboard" Hot 100 and sold millions in the United States, making him \"the most successful male UK urban artist in US chart history\" at the time\dots} \\
        & \textbf{Jaccard Similarity}: 0.16 & \textbf{Jaccard Similarity}: 0.14 \\
        & \textbf{BertScore}: 0.85 & \textbf{BertScore}: 0.86 \\
        \midrule
        Model output & Karachi & Elton John \\
        \bottomrule
    \end{tabular}
    \end{adjustbox}
    \caption{Some examples where both the generator and reader are GPT-3.5. We highlight the incorrect candidate answers in the context in \sethlcolor{pink}\hl{pink}, and the correct answers in the context in \sethlcolor{green}\hl{green}.}
    \label{case_similarity}
\end{table*}

\subsection{The Effectiveness of Exactly Matching}
\Cref{others_ratio} demonstrates that the proportion of ``Others'' is significantly lower relative to the disparities between ``Gen'' and ``Ret'', its impact on the conclusions of this paper is negligible.
\begin{table*}[t]
    \centering
    \begin{adjustbox}{max width=\textwidth}
    \begin{tabular}{llcccccccccccc}
        \toprule
    	\multirow{2}{*}{Reader} &  \multirow{2}{*}{Generator} &  \multicolumn{3}{c}{NQ-AIG} & \multicolumn{3}{c}{NQ-AIR} & \multicolumn{3}{c}{TQA-AIG} & \multicolumn{3}{c}{TQA-AIR} \\
        \cmidrule(l){3-5}  \cmidrule(l){6-8} \cmidrule(l){9-11} \cmidrule(l){12-14}
        & & Gen & Ret & Others & Gen & Ret & Others & Gen & Ret & Others & Gen & Ret & Others \\
    	\midrule
        GPT 4 & GPT 4 & 0.9125 & 0.0589 & 0.0286 & 0.7379 & 0.1743 & 0.0878 & 0.9387 & 0.0304 & 0.031 & 0.7651 & 0.1762 & 0.0587 \\
        GPT 3.5 & GPT 3.5 & 0.9177 & 0.0449 & 0.0374 & 0.7083 & 0.1470 & 0.1447 & 0.9347 & 0.026 & 0.0393 & 0.7332 & 0.1775 & 0.0893 \\
        Llama2-13b & Llama2-13b & 0.8966 & 0.0500 & 0.0534 & 0.7216 & 0.1811 & 0.0973 & 0.9071 & 0.0433 & 0.0496 & 0.7212 & 0.1918 & 0.0870 \\
        Llama2-7b & Llama2-7b & 0.7041 & 0.1082 & 0.1876 & 0.6148 & 0.2071 & 0.1781 & 0.7973 & 0.0927 & 0.1100 & 0.6555 & 0.2101 & 0.1344 \\
    	\bottomrule
    \end{tabular}
    \end{adjustbox}
    \caption{``Gen'' denotes the proportion of responses that match the candidate answer within generated contexts, whereas ``Ret'' refers to the proportion of matching the candidate answer within retrieved contexts. ``Others'' encompasses the proportion of responses that do not align with either category. \textit{Given that the proportion of ``Others'' is significantly lower relative to the disparities between ``Gen'' and ``Ret'', its impact on the conclusions of this paper is negligible.}}
    \label{others_ratio}
\end{table*}

\end{document}